%% file: main.tex
\documentclass[journal ]{new-aiaa}
\usepackage[utf8]{inputenc}
\usepackage{textcomp}

\usepackage{graphicx}
\usepackage{amsmath}
\usepackage[version=4]{mhchem}
\usepackage{siunitx}
\usepackage{longtable,tabularx}
\usepackage{multirow}
\usepackage{makecell}

\title{Remarks on stochastic cloning and delayed-state filtering}

\author{Tara Mina \footnote{Research Engineer II, Daniel Guggenheim School of Aerospace Engineering.}, Lindsey Marinello \footnote{Graduate Student, Daniel Guggenheim School of Aerospace Engineering.}, and John Christian \footnote{Associate Professor and Associate Chair for Graduate Programs, Daniel Guggenheim School of Aerospace Engineering, Associate Fellow AIAA.}}
\affil{Georgia Institute of Technology, Atlanta, Georgia, 30332}

\input{macros}

\begin{document}

\maketitle

\input{sections/0_abstract}
\input{sections/1_intro}

\input{sections/2_historyAndLexicon}
\input{sections/3_traditionalKF}
\input{sections/4_delayedState}
\input{sections/5_stochasticCloning}

\input{sections/6_dskf}
\input{sections/7_equivalence}
\input{sections/8_fewerClonedStates}
\input{sections/9_computationsMemory}

\input{sections/10_conclusion}
\input{sections/11_appendix}
\input{sections/12_fundingAcknowledgments}

\bibliography{references}

\end{document}

%% file: macros.tex
\newcommand{\trans}{^\top}
\newcommand{\tr}[1]{\text{tr}\left(#1\right)}
\newcommand{\inv}{^{-1}}
\newcommand{\prob}[1]{\mathbb{P}\left\{#1\right\}}
\newcommand{\expec}[1]{\mathbb{E}\left[#1\right]}
\newcommand{\stochproc}[2]{(#1_{#2} : #2 \in \stochproctimeset )}
\newcommand{\stochproctimeset}{\mathbb{T}}

\newcommand{\state}{x}

\newcommand{\eststateerr}{\delta \est{x}}
\newcommand{\action}{u}
\newcommand{\meas}{\tilde{y}}
\newcommand{\procnoise}{w}
\newcommand{\measnoise}{v}
\newcommand{\statecov}{P}
\newcommand{\actionmat}{B}
\newcommand{\procnoisemat}{G}
\newcommand{\unscaledprocnoisecov}{Q}
\newcommand{\procnoisecov}{S}
\newcommand{\measnoisecov}{R}
\newcommand{\kalmangain}{K}
\newcommand{\numstates}{n}
\newcommand{\numprior}{\ell}
\newcommand{\nummeas}{m}
\newcommand{\innov}{\delta \tilde{y}}
\newcommand{\innovcov}{\Upsilon}
\newcommand{\Mcoeff}{\corr{M}}
\newcommand{\Lcoeff}{\corr{L}}
\newcommand{\previdx}{j}
\newcommand{\curidx}{k}

\newcommand{\pos}{p}
\newcommand{\aug}[1]{\breve{#1}}
\newcommand{\corr}[1]{\mathcal{#1}}

\newcommand{\fwdTidx}[2]{{\corr{#1}^\to\tidx{#2}}}
\newcommand{\bwdTidx}[2]{{\corr{#1}^\gets\tidx{#2}}}
\newcommand{\linmeasmdlprev}{J}
\newcommand{\crosscov}{N}

\newcommand{\tidx}[1]{_{#1}}

\newcommand{\ellPriorState}{{\state}\tidx{\previdx}^{(\numprior)}}
\newcommand{\ellPriorStateEst}{{\est{\state}}\tidx{\previdx}^{(\numprior)}}
\newcommand{\ellPriorStateCov}{{\statecov}\tidx{\previdx}^{(\numprior)}}
\newcommand{\ellPriorStateCovCol}{{\statecov}^{(:,\numprior)}\tidx{\previdx}}
\newcommand{\ellMeasMdl}{{\linmeasmdl}\tidx{\previdx,\curidx}^{(\numprior)}}
\newcommand{\ellProjMat}{\Pi_\numprior}
\newcommand{\ellAugH}{\aug{\linmeasmdl}\tidx{\curidx}^{(\numprior)}}
\newcommand{\ellSmat}{\Smat^{(\numprior)}}
\newcommand{\ellaugPredState}{{\aug{\est{\state}}}\tidx{\curidx}^{-(\numprior)}}
\newcommand{\ellaugPredStateCov}{{\aug{\statecov}}\tidx{\curidx}^{-(\numprior)}}
\newcommand{\ellSTMInv}{\stm\tidx{\previdx,\curidx}^{(\numprior)}}

\newcommand{\linmeasmdl}{H}
\newcommand{\stm}{\Phi}
\newcommand{\gaussian}[2]{\mathcal{N}(#1, #2)}

\newcommand{\est}[1]{\hat{#1}}
\newcommand{\upd}[1]{{#1}^{+}}
\newcommand{\pred}[1]{{#1}^{-}}
\newcommand{\timeidx}[1]{t_{#1}}

\newcommand{\eye}{\mathbb{I}}
\newcommand{\eyedim}[1]{\eye_{#1}}
\newcommand{\zeromat}{\mathbf{0}}
\newcommand{\zeromatdim}[1]{\zeromat_{#1}}
\newcommand{\Smat}{U}  

\newcommand{\Rdim}[1]{\mathbb{R}^{#1}}

%% file: sections/0_abstract.tex
\begin{abstract}
Many estimation problems in aerospace navigation and robotics involve measurements that depend on prior states. 
A prominent example is odometry, which measures the relative change between states over time.
Accurately handling these \emph{delayed-state measurements} requires capturing their correlations with prior state estimates, and a widely used approach is stochastic cloning~(SC), which augments the state vector to account for these correlations. 
This work revisits a long-established but often overlooked alternative---the delayed-state Kalman filter---and demonstrates that a properly derived filter yields exactly the same state and covariance update as SC, without requiring state augmentation. 
Moreover, two equivalent formulations of the delayed-state Kalman filter~(DSKF) are presented, providing complementary perspectives on how the prior-state measurement correlations can be handled within the generalized Kalman filter. 
These formulations are shown to be comparable to SC in asymptotic computational and memory complexity, while one DSKF formulation can offer reduced arithmetic and storage costs for certain problem dimensions. 
Our findings clarify a common misconception that Kalman filter variants are inherently unable to handle correlated delayed-state measurements, demonstrating that an alternative formulation achieves the same results without state augmentation.
\end{abstract}

%% file: sections/1_intro.tex
\section{Introduction}
\lettrine{M}{any} state estimation problems involve measurements that depend on prior states.
This dependency is oftentimes either due to processing delays or because the measurement itself depends upon state information over multiple time steps, such as with odometry measurements. 
Such \textit{delayed-state measurements} further introduce correlations between the measurement error and prior state estimate error, violating a common assumption in many Kalman filter formulations—--namely, that measurements are conditionally independent of prior states given the current state~\cite{Maybeck:1982,thrun2005probabilistic}. 
This issue is not due to a fundamental limitation of the Kalman filter itself, but rather of its conventional formulation.
Accurate handling of these measurements requires capturing their prior-state correlations, and this work aims to bring new clarity to the various approaches used to solve this problem and the relationships between them.

In aerospace and robotics systems, delayed-state measurements often arise in the context of odometry, where observations describe how a vehicle’s pose changes over time.  
Odometry-based estimation is ubiquitous in modern robotics, with numerous works discussing topics of~visual-inertial odometry~(VIO)~\cite{Nister:2004,Scaramuzza:2011}, LiDAR-inertial odometry~(LIO)~\cite{shan2020lio,xu2021fast}, and wheel odometry~\cite{chenavier1992position,borenstein1996measurement,welte2019four}. 
These techniques have seen widespread adoption on ground and aerial robotic platforms~\cite{nister2006visual,scaramuzza2019visual,mohamed2019survey} as well as in space exploration settings~\cite{helmick2004path,cheng2004mars,cheng2005visual,Christian:2021vo,christian2025optical,Bayard:2019}.
Odometry measurements can be used within a simultaneous localization and mapping (SLAM) framework or in a map-free approach. 
In situations where real-time map construction is not important, the map-free approach is sometimes attractive due to its reduced computational complexity.

Proper handling of delayed-state measurements, including odometry, can be achieved via sequential filtering (e.g., delayed-state Kalman filter~\cite{brown1997kalman}, multi-state constraint Kalman filter~\cite{mourikis2007multi}), Kalman smoothing~\cite{rauch1963solutions,brown1997kalman}, fixed-lag smoothing~\cite{challa2002fixed}, dynamic Bayesian networks~\cite{ranganathan2007fast}, factor graphs~\cite{Forster:2017}, or related methods. 
This work focuses on processing these measurements within the sequential filter framework. 
However, the framework based on Refs.~\cite{brown1968kalman,brown1997kalman} and presented in Section~\ref{sec:brownAndHwangFilter} can be extended to Kalman smoothing, including the fixed-lag case~\cite{finch1970smoothing}.

Delayed-state filtering is also relevant in radio navigation applications.
In particular, delta-range GNSS measurements, oftentimes referred to as time-differenced carrier phase~(TDCP) measurements, are a form of delayed-state measurement obtained by counting the total carrier phase cycles received from a GNSS satellite signal across two time steps~\cite{brown1987proper,kaplan2017understanding}.
These measurements are analogous to a change in range to the satellite. 
Proper handling of TDCP requires accounting for state dependencies across time steps, particularly in precise navigation applications~\cite{brown1987proper, kaplan2017understanding, guillard2022using}.

Stochastic cloning (SC)~\cite{roumeliotis2002stochastic} is a widely accepted approach to handle the correlations introduced by delayed-state measurements, whereby prior states are appended to the state vector and the covariance is augmented to capture the corresponding cross-correlations.
As discussed in Section~\ref{ssec:lexicon}, this general concept appears under various names within different filtering applications and frameworks, including the multi-state constraint Kalman filter (MSCKF)~\cite{mourikis2007multi}, and has been applied across a wide range of domains. 
In space applications, SC has been used to accommodate delayed measurements in the Mars 2020 Lander Vision System~(LVS)~\cite{Johnson:2023} and for processing odometry measurements on the NASA Mars Ingenuity helicopter~\cite{Bayard:2019}. 
It has also been proposed for precise navigation using TDCP GNSS measurements, both for ground vehicles~\cite{chiang2020design} and lunar orbiters~\cite{iiyama2023terrestrial}. 
SC has also been proposed for visual odometry in wearable devices, such as augmented reality binoculars~\cite{oskiper2015augmented}, foot-mounted inertial units for pedestrian step detection~\cite{ruppelt2015novel}, and assistive navigation technologies for visually impaired individuals~\cite{schwarze2015intuitive}.

Beyond navigation applications, delayed-state measurements also arise in other domains such as chemical and biochemical process systems~\cite{gopalakrishnan2011incorporating}, where measurements of past states may arrive with latency or out of sequence. 
In these cases, consistency in estimation is commonly maintained by augmenting the current state with past states~\cite{gopalakrishnan2011incorporating}—an approach analogous to stochastic cloning when using a single delayed state~\cite{roumeliotis2002stochastic}.

A prevailing argument for using stochastic cloning, or state augmentation, is that the standard Kalman filter cannot properly handle the correlations introduced by delayed or relative-state measurements. 
This view has been stated or implied in hundreds of papers over the past two decades. 
Contrary to this popular assertion, however, this work demonstrates that a correctly derived Kalman filter~\cite{brown1968kalman}, also sometimes called the delayed-state Kalman filter~(DSKF)~\cite{brown1997kalman,brown1987proper}, produces the exact same estimated state and covariance as stochastic cloning. 
This work presents two equivalent formulations of the DSKF, which offer complementary perspectives on how these correlations can be handled within the generalized Kalman filter.
Additionally, the computational complexity and memory usage are examined for stochastic cloning and both DSKF formulations.
While arithmetic and memory performance is highly dependent on implementation details, our analysis---based on the presented filter formulations and efficient computation sequences for the filter updates---finds that the various methods are generally comparable in asymptotic order.
Within this shared order, however, one of the DSKF formulations can exhibit lower arithmetic and memory usage than stochastic cloning in certain cases.

The remainder of the paper is organized as follows. 
Section~\ref{ssec:lexicon} provides a brief overview on the history of delayed-state filtering and clarifies the terminology that has emerged in the literature to describe oftentimes similar, or even equivalent, techniques.
Section~\ref{sec:kf} reviews the traditional Kalman filter and establishes the notation used throughout the paper, while Section~\ref{sec:delayedStateDep} illustrates the challenges posed by delayed-state measurement dependence for this standard framework.
Sections~\ref{sec:sc}~and~\ref{sec:brownAndHwangFilter} introduce two high-level frameworks for handling the delayed-state measurement scenario: stochastic cloning and the delayed-state Kalman filter, respectively. 
Section~\ref{sec:brownAndHwangFilter} further presents two equivalent formulations of the delayed-state Kalman filter. 
The equivalence between these two popular frameworks is demonstrated in Section~\ref{sec:equivalence}.
Section~\ref{sec:simplifications} introduces simplifications to these filter formulations by exploiting sparse prior-state dependence in the measurement. 
Finally, Section~\ref{sec:compAndMemory} compares the computational and memory requirements of the presented filtering frameworks, and Section~\ref{sec:conc} concludes the paper.

%% file: sections/2_historyAndLexicon.tex
\section{The History and Lexicon of Delayed-State Filtering}\label{ssec:lexicon}
Over the years, various terms have been introduced to describe similar---or, oftentimes, equivalent---techniques in delayed-state filtering. 
This usually happens by accident rather than on purpose.
Unfortunately, however, the introduction of different terms to describe the same idea creates confusion, misattribution of discovery, and makes it more difficult to locate relevant prior works. 
With an intent to clarify the relationships between similar ideas, this section provides (1)~an overview of the most common terms and (2)~a summary of some important historical connections.

To the best of the authors' knowledge, the earliest formulation of a Kalman filter variant designed to handle delayed-state measurements appeared in the work of Brown and Hartman~(1968)~\cite{brown1968kalman}.
Their derivation introduces modified update equations for the recursive Kalman filter that explicitly account for correlations between the measurement and the prior state.
Although the 1968 paper did not assign a specific name to this generalized Kalman filter, Brown later refers to this filter as the \textit{delayed-state Kalman filter}~\cite{brown1987proper} in the context of GPS delta-range measurement processing and in his textbook on Kalman filtering, co-authored with Hwang~\cite{brown1997kalman}.

Alternatively, one can augment the Kalman filter's state vector by including the prior state, thereby directly representing the measurement model as a function of the augmented state vector.
This approach preserves the familiar Kalman filter structure, which is more commonly recognized.
This technique is popularly termed as \textit{stochastic cloning}, coined by Roumeliotis and Burdick~(2002)~\cite{roumeliotis2002stochastic}.

However, a deeper examination of the literature reveals that the stochastic cloning approach predates the term.
In fact, Brown and Hartman (1968) describe a method of “twisting the model” for the delayed-state measurement scenario in order to conform to the standard Kalman filter formulation, arriving at the same mathematical expressions that were later proposed for stochastic cloning~\cite{roumeliotis2002stochastic}.
At the time, Brown and Hartman (1968) refer to this as the ``double-state approach''~\cite{brown1968kalman}, but then ultimately focus on deriving a recursive update form for the delayed-state Kalman filter using the original state vector.
While they note that both formulations should yield equivalent results, this equivalence was not explicitly demonstrated, as it was not the focus of their work and analysis—an observation that motivates a more explicit comparison in the present work.

In the literature, two related contexts arise in the delayed-state measurement problem.
First, a measurement may depend only on a single prior state but is received or processed at a later time. This occurs frequently in practice for real-time onboard filtering, as measurements are inherently delayed relative to the current estimation time.
Second, a measurement may simultaneously depend on multiple states at different times — for instance, in odometry measurements that relate states across time.
The first scenario is a special case of the second, more general formulation.
Both the delayed-state Kalman filter (DSKF) and stochastic cloning (SC) have been used in prior works to address both the specific case of delayed measurements and the more general case of multi-state measurements.

Various terms have been used in the literature to describe techniques equivalent to \textit{stochastic cloning}, where the state vector is augmented with prior states to address the delayed-state measurement problem. 
One related term is the \textit{stochastic cloning Kalman filter (SC-KF)}~\cite{mourikis2007sc}. 
The \textit{multi-state constraint Kalman filter (MSCKF)}~\cite{mourikis2007multi} similarly augments the filter's state and covariance matrix with copies of prior states and associated covariances at the times when camera measurements were taken. 
An equivalent framework to the MSCKF has also been referred to as the \textit{stochastic cloning sliding-window EKF}~\cite{liang2021novel}.
Related approaches involving cloning and augmentation of prior states have also been referred to in prior works as \textit{latency compensation}~\cite{merwe2004sigma}, the \textit{augmented state extended Kalman filter (AS-EKF)}~\cite{das2019ekf}, and \textit{reduced-order Kalman filtering with relative measurements}~\cite{bayard2009reduced}.
This list is not meant to be exhaustive, but rather to illustrate the range of terminologies that effectively describe the same underlying method.

A related term that is used in the context of delayed measurements is \textit{measurement extrapolation}~\cite{larsen1998incorporation} or \textit{Larsen's method}~\cite{silvestrini2022optical}. 
This approach derives a linear expression that maps a delayed measurement—dependent on a past state—to the current time step, thereby relating the past measurement to the current state and motivating the term ``extrapolation.''
The resulting correlation between measurement noise and state uncertainty is then explicitly modeled and incorporated in the Kalman filter update. 
Similar to the delayed-state Kalman filter, this approach handles the resulting correlations directly within the filter equations, rather than through state augmentation. 
While specifically designed for delayed measurements, this approach can also be framed within the more general delayed-state Kalman filter formulation~\cite{brown1997kalman}.

Given the variety of terminology that has been used for closely related techniques, this work adopts the widely used term \textit{stochastic cloning} to refer to the augmentation of the state vector with prior states. 
Brown’s terminology, \textit{delayed-state Kalman filter}, is used for the generalized Kalman filter formulation that explicitly accounts for correlations between delayed-state measurements and the prior state.

%% file: sections/3_traditionalKF.tex
\section{Traditional Kalman Filter}\label{sec:kf}

\subsection{Dynamics and Observation Models}
For a linear, discrete Kalman filter~\cite{carpenter2018navigation,brown1997kalman}, suppose linear dynamics propagate the state at time $\timeidx{\previdx}$ to time $\timeidx{\curidx} > \timeidx{\previdx}$
\begin{align}
    \state\tidx{\curidx} &= \stm\tidx{\curidx,\previdx} \state\tidx{\previdx} + \actionmat\tidx{\curidx} \action\tidx{\curidx} + \procnoisemat\tidx{\curidx} \procnoise\tidx{\curidx} , \label{eq:lindyn}
\end{align}
where $\state\tidx{\curidx}\in\Rdim{\numstates}$ denotes the state at time $\timeidx{\curidx}$, $\action\tidx{\curidx}$ is the action, and $\procnoise\tidx{\curidx}\sim\gaussian{0}{\unscaledprocnoisecov\tidx{\curidx}}$ is the Gaussian process noise with covariance $\unscaledprocnoisecov\tidx{\curidx}$.
$\stm\tidx{\curidx,\previdx}$ is the state transition matrix~(STM) from $\timeidx{\previdx}$ to $\timeidx{\curidx}$, $\actionmat\tidx{\curidx}$ is the control-input model, and $\procnoisemat\tidx{\curidx}$ is the noise coupling matrix, mapping the process noise to its effect in the state.

Observations are modeled as a linear function of the state
\begin{align}
    \meas\tidx{\curidx} &= \linmeasmdl\tidx{\curidx} \state\tidx{\curidx} + \measnoise\tidx{\curidx} \in\Rdim{\nummeas}, \label{eq:measmdl}
\end{align}
where $\linmeasmdl\tidx{\curidx}$ is the linear observation model and $\measnoise\tidx{\curidx}\sim\gaussian{0}{\measnoisecov\tidx{\curidx}}$ is the Gaussian measurement noise with covariance $\measnoisecov\tidx{\curidx}$.
\subsection{Predict and Update Steps}
Given the state estimate and covariance at $\previdx<\curidx$, the predicted state and covariance at $\timeidx{\curidx}$ is evaluated as
\begin{gather}
    \pred{\est{\state}}\tidx{\curidx} = \stm\tidx{\curidx,\previdx} \upd{\est{\state}}\tidx{\previdx} + \actionmat\tidx{\curidx} \action\tidx{\curidx}, \label{eq:predStateEstimate} \\
    \pred{\statecov}\tidx{\curidx} = \stm\tidx{\curidx,\previdx}\upd{\statecov}\tidx{\previdx}{\stm\trans\tidx{\curidx,\previdx}} + \procnoisecov\tidx{\curidx}, \label{eq:predCov_kf}
\end{gather}
where 
{ $\upd{\statecov}\tidx{\previdx}$  is the covariance of the prior estimate $\upd{\est{\state}}\tidx{\previdx}$,} 
the superscript $\upd{}$ denotes quantities updated with the corresponding observation at the given time step, the superscript $\pred{}$ signifies terms predicted forward in time, and the process noise covariance is
\begin{align}
    \procnoisecov\tidx{\curidx} \triangleq \procnoisemat\tidx{\curidx} \unscaledprocnoisecov\tidx{\curidx} \procnoisemat\trans\tidx{\curidx}. \label{eq:procnoisecov}
\end{align}

After obtaining the set of observations $\meas\tidx{\curidx}$, the Kalman filter evaluates the measurement innovation $\innov\tidx{\curidx}$ and covariance $\innovcov\tidx{\curidx}$
\begin{gather}
    \innov\tidx{\curidx} = \meas\tidx{\curidx} - \linmeasmdl\tidx{\curidx}\pred{\est{\state}}\tidx{\curidx}, \\
    \innovcov\tidx{\curidx} = \linmeasmdl\tidx{\curidx}\pred{\statecov}\tidx{\curidx} \linmeasmdl\trans\tidx{\curidx} + \measnoisecov\tidx{\curidx},
\end{gather}
then maps this information to the state space via the Kalman gain $\kalmangain\tidx{\curidx}$ in order to evaluate the updated state and covariance
\begin{gather}
    \kalmangain\tidx{\curidx} = \pred{\statecov}\tidx{\curidx}\linmeasmdl\trans\tidx{\curidx} \innovcov\inv\tidx{\curidx}, \\
    \upd{\est{\state}}\tidx{\curidx} =  \pred{\est{\state}}\tidx{\curidx} + \kalmangain\tidx{\curidx} \innov\tidx{\curidx}, \\
    \upd{\statecov}\tidx{\curidx} = (\eye - \kalmangain\tidx{\curidx}\linmeasmdl\tidx{\curidx})\pred{\statecov}\tidx{\curidx}(\eye - \kalmangain\tidx{\curidx}\linmeasmdl\tidx{\curidx})\trans + \kalmangain\tidx{\curidx}\measnoisecov\tidx{\curidx}\kalmangain\trans\tidx{\curidx},
\end{gather}
where $\eye$ denotes the identity matrix.

%% file: sections/4_delayedState.tex
\section{Delayed-State Dependence in Sequential Filters}\label{sec:delayedStateDep}
A fundamental assumption in conventional sequential filtering is that the state follows a \textit{Markov process}~\cite{hajek2015random}.
In other words, the random process $\stochproc{\state}{\curidx}$ describing the state over the set of times $\stochproctimeset$ satisfies the \textit{Markov property}, allowing for the conditional independence relation
\begin{align}
    \prob{\state\tidx{\curidx+1} | \state\tidx{\curidx}, \state\tidx{\curidx-1}, \cdots, \state\tidx{0}} = \prob{\state\tidx{\curidx+1} | \state\tidx{\curidx}}.
\end{align}
Sequential filters oftentimes assume the observations $\stochproc{\meas}{\curidx}$ follow a \textit{hidden Markov model}~(HMM) with respect to the state, satisfying a similar conditional independence relation
\begin{align}
    \prob{\meas\tidx{\curidx} | \state\tidx{\curidx}, \state\tidx{\curidx-1}, \cdots, \state\tidx{0}} = \prob{\meas\tidx{\curidx} | \state\tidx{\curidx}}.\label{eq:probObsMdl}
\end{align}
Indeed, the Kalman filter observation model in Eq.~\eqref{eq:measmdl} is only a function of the current state,
which inherently follows a HMM, due to its conditional independence from prior states, given the current state.

A delayed-state dependency poses a problem for sequential filtering, since it breaks the HMM assumption.
For example, LiDAR odometry provides relative motion measurement information between two positions at different times, $\timeidx{\curidx}$~and~$\timeidx{\previdx}$
\begin{align}
    \meas\tidx{\curidx} = (\pos\tidx{\curidx} - \pos\tidx{\previdx}) + \measnoise\tidx{k},\label{eq:vomeas}
\end{align}
with $\pos\tidx{\curidx}$ denoting the position state at time $\timeidx{\curidx}$.
As written, the observation model $\stochproc{\state}{\curidx}$ no longer follows a HMM.

%% file: sections/5_stochasticCloning.tex
\section{Stochastic Cloning Kalman Filter}\label{sec:sc}
Stochastic cloning (SC)~\cite{roumeliotis2002stochastic} handles delayed-state dependencies by augmenting the state vector to regain Markovity
\begin{align}
    \aug{\state}\tidx{\curidx} \triangleq \begin{bmatrix} \state\tidx{\previdx}\trans & \state\tidx{\curidx}\trans \end{bmatrix}\trans. \label{eq:augState}
\end{align}
While it is possible to augment the state with prior states from multiple times—and the extension to an arbitrary number of times is straightforward—for clarity, this work focuses on the case involving only two time instances.

The linear measurement with delayed-state dependence can be modeled with respect to the augmented state in Eq.~\eqref{eq:augState} as
\begin{align}
    \meas\tidx{k} &= \aug{\linmeasmdl}\tidx{k} \aug{\state}\tidx{k} + \measnoise\tidx{k}.\label{eq:augMeasMdl} 
\end{align}
Indeed, for the LiDAR odometry case in Eq.~\eqref{eq:vomeas}, letting ${\state}\tidx{k} = \pos\tidx{k}$ enforces the augmented measurement matrix to be $\aug{\linmeasmdl}\tidx{k} = \begin{bmatrix} -\eye & \eye\end{bmatrix}$.
Thus, by re-defining the state, the observation model regains the property of following a HMM with respect to the augmented state.

\subsection{Predict Step}
For the predict step, the augmented state estimate becomes
\begin{align}
     \pred{\aug{\est{\state}}}\tidx{\curidx} = \begin{bmatrix} \upd{\est{\state}}\tidx{\previdx} \\ \pred{\est{\state}}\tidx{\curidx} \end{bmatrix} \in \Rdim{2\numstates} , \label{eq:augStateDef}
\end{align}
thereby concatenating (or ``cloning'') the previous state estimate at time~$\timeidx{\previdx}$ with the original predicted state estimate.
The dynamics for the augmented system can be represented as
\begin{align}
    \pred{\aug{\est{\state}}}\tidx{\curidx} &= \begin{bmatrix} \zeromatdim{\numstates} & \eyedim{\numstates} \\ \zeromatdim{\numstates} &  \stm\tidx{\curidx,\previdx} \end{bmatrix} \upd{\aug{\est{\state}}}\tidx{\previdx} + \begin{bmatrix} 0 \\ \actionmat\tidx{k} \end{bmatrix} \action\tidx{\curidx} , \label{eq:augPredictFullMat} \\
    &\triangleq \aug{\stm}\tidx{\curidx,\previdx} \upd{\aug{\est{\state}}}\tidx{\previdx} + \aug{\actionmat}\tidx{\curidx} \action\tidx{\curidx},
\end{align}
where $\zeromatdim{\numstates}$ is an ${\numstates \times \numstates}$ matrix of zeros and $\eyedim{\numstates}$ is the  ${\numstates \times \numstates}$ identity matrix.
The augmented predicted covariance is
\begin{align}
     \pred{\aug{\statecov}}\tidx{k} &= \begin{bmatrix}
        \upd{\statecov}\tidx{\previdx} & \upd{\statecov}\tidx{\previdx} \stm\trans\tidx{\curidx,\previdx}  \\
        \stm\tidx{\curidx,\previdx} \upd{\statecov}\tidx{\previdx} & \pred{\statecov}\tidx{\curidx} 
     \end{bmatrix} , \label{eq:augCovDef}
\end{align}
The covariance prediction can also be represented in the traditional Kalman filtering formulation as
\begin{gather} 
     \pred{\aug{\statecov}}\tidx{\curidx} = \aug{\stm}\tidx{\curidx,\previdx} \upd{\aug{\statecov}}\tidx{\previdx}\aug{\stm}\trans\tidx{\curidx,\previdx} + \aug{\procnoisecov}\tidx{\curidx}, \\
     \aug{\procnoisecov}\tidx{\curidx} = \begin{bmatrix}
        \zeromatdim{\numstates} & \zeromatdim{\numstates} \\
       \zeromatdim{\numstates} & \procnoisecov\tidx{\curidx} 
     \end{bmatrix} .
\end{gather}

\subsection{Update Step}
With the augmented state, the update step can be performed in an identical manner as in the traditional Kalman filter using the delayed-state-dependent measurement model in Eq.~\eqref{eq:augMeasMdl}.
The innovation is computed with the augmented state as
\begin{gather}
    {\innov}\tidx{\curidx} = \meas\tidx{\curidx} - \aug{\linmeasmdl}\tidx{\curidx} \pred{\aug{\est{\state}}}\tidx{\curidx}, \label{eq:augMatInnovation} \\
    {\innovcov}\tidx{\curidx} = \aug{\linmeasmdl}\tidx{\curidx} \pred{\aug{\statecov}}\tidx{\curidx} \aug{\linmeasmdl}\trans\tidx{\curidx} + \measnoisecov\tidx{\curidx}, \label{eq:scInnovCov}
\end{gather}
then mapped to the state space for the Kalman update as
\begin{gather}
    \aug{\kalmangain}\tidx{\curidx} = \pred{\aug{\statecov}}\tidx{\curidx} \aug{\linmeasmdl}\trans\tidx{\curidx} {\innovcov}\inv\tidx{\curidx}, \label{eq:sckalmangain}\\
    \upd{\aug{\est{\state}}}\tidx{\curidx} =  \pred{\aug{\est{\state}}}\tidx{\curidx} + \aug{\kalmangain}\tidx{\curidx} {\innov}\tidx{\curidx}, \\
    \upd{\aug{\statecov}}\tidx{\curidx} = (\eye - \aug{\kalmangain}\tidx{\curidx} \aug{\linmeasmdl}\tidx{\curidx}) \pred{\aug{\statecov}}\tidx{\curidx} (\eye - \aug{\kalmangain}\tidx{\curidx}\aug{\linmeasmdl}\tidx{\curidx})\trans + \aug{\kalmangain}\tidx{\curidx}\measnoisecov\tidx{\curidx}\aug{\kalmangain}\trans\tidx{\curidx}. \label{eq:stateUpdateSC}
\end{gather}
Note that the update step acts on the complete augmented state.
However, from Eq.~\eqref{eq:augPredictFullMat}, the cloned state will be discarded in the next predict step.
Thus, to reduce computational cost, only the $\numstates \times \nummeas$ lower submatrix of the Kalman gain associated with the current state needs to be computed. 
Additionally, only the last $\numstates$ elements of the augmented vector, corresponding to the original state of interest, needs to be updated
\begin{gather}
     \aug{\kalmangain}\tidx{\curidx} = \begin{bmatrix} \cdot &  \aug{\kalmangain}\tidx{\curidx,\curidx}\trans \end{bmatrix}\trans, \label{eq:scAugKalmanGain}\\
     \upd{\est{\state}}\tidx{\curidx} =  \pred{\est{\state}}\tidx{\curidx} + \aug{\kalmangain}\tidx{\curidx,\curidx} {\innov}\tidx{\curidx}. \label{eq:scStateUpdate}
\end{gather}
Similarly, only the $\numstates \times \numstates$ covariance corresponding to the state of interest is required. 
Operating on only the relevant submatrices in Eq.~\eqref{eq:stateUpdateSC} yields the simplified covariance update 
\begin{gather}
    \upd{\aug{\statecov}}\tidx{\curidx,\curidx} = \left( \Smat - \aug{\kalmangain}\tidx{\curidx,\curidx} \aug{\linmeasmdl}\tidx{\curidx} \right) \pred{\aug{\statecov}}\tidx{\curidx}\left( \Smat - \aug{\kalmangain}\tidx{\curidx,\curidx} \aug{\linmeasmdl}\tidx{\curidx} \right)\trans +  \aug{\kalmangain}\tidx{\curidx,\curidx} \measnoisecov\tidx{\curidx} \aug{\kalmangain}\tidx{\curidx,\curidx}\trans, \label{eq:scCovFinal}
\end{gather}
where $\Smat \triangleq \begin{bmatrix} \zeromatdim{\numstates} & \eyedim{\numstates} \end{bmatrix}$.
Note that not all implementations of stochastic cloning incorporate these reduced submatrix operations in the update step (e.g., the update may be performed on the full augmented state and covariance before extracting the relevant subcomponents). 
Accordingly, any associated computational and memory reductions are realized only when the update is implemented with these reduced submatrix operations.

%% file: sections/6_dskf.tex
\section{Delayed-State Kalman Filter}\label{sec:brownAndHwangFilter}
The delayed-state dependency can also be handled by explicitly modeling the correlation between the measurement and prior state.
Traditional Kalman filters assume these are conditionally uncorrelated given the current state, implying uncorrelated process and measurement noise at time~$\timeidx{\curidx}$.
This section outlines the derivation of a generalized version of the Kalman filter~\cite{brown1997kalman}, called the \textit{delayed-state Kalman filter}~(DSKF), which introduces additional terms in the Kalman gain and covariance update to account for these dependencies.
Two equivalent formulations of the DSKF arise from different choices of state propagation direction between the prior state at~$t_j$ and the current state at~$t_k$. This equivalence is demonstrated in the Appendix.

To begin, let us use Eq.~\eqref{eq:augMeasMdl} to define our measurement model
\begin{align}
     \meas\tidx{\curidx} =  \begin{bmatrix} {\linmeasmdl}\tidx{\previdx,\curidx} & {\linmeasmdl}\tidx{\curidx,\curidx}\end{bmatrix}  \begin{bmatrix} {{\state}}\tidx{\previdx} \\ {{\state}}\tidx{\curidx}\end{bmatrix} + \measnoise\tidx{\curidx},\label{eq:augTrueMeas}
\end{align}
where ${\linmeasmdl}\tidx{\previdx,\curidx}$ is the measurement matrix coefficient at time~$\timeidx{\curidx}$ with respect to the prior state at time~$\timeidx{\previdx}$.
Note that $\begin{bmatrix} {\linmeasmdl}\tidx{\previdx,\curidx} & {\linmeasmdl}\tidx{\curidx,\curidx}\end{bmatrix} =  \aug{\linmeasmdl}\tidx{k}$, and thus the measurement model in Eq.~\eqref{eq:augTrueMeas} is equivalent to Eq.~\eqref{eq:augMeasMdl}. The corresponding innovation becomes
\begin{align}
     \innov\tidx{\curidx} = \meas\tidx{\curidx} - \begin{bmatrix} {\linmeasmdl}\tidx{\previdx,\curidx} & {\linmeasmdl}\tidx{\curidx,\curidx}\end{bmatrix}  \begin{bmatrix} \upd{\est{\state}}\tidx{\previdx} \\ \pred{\est{\state}}\tidx{\curidx}\end{bmatrix}.\label{eq:delayedStateMeas}
\end{align}

Rather than concatenating the state at two times as in SC, the DSKF aims to formulate the problem while retaining the state at only a single time. 
Through the STM, the states $\upd{\est{\state}}\tidx{\previdx}$ and $\pred{\est{\state}}\tidx{\curidx}$ are related via the bidirectional mapping
\begin{align}
    \pred{\est{\state}}\tidx{\curidx} &= \stm\tidx{\curidx,\previdx} \upd{\est{\state}}\tidx{\previdx} + \actionmat\tidx{\curidx} \action\tidx{\curidx}, \\
    \upd{\est{\state}}\tidx{\previdx} &= \stm\tidx{\previdx,\curidx}\left(\pred{\est{\state}}\tidx{\curidx} - \actionmat\tidx{\curidx} \action\tidx{\curidx} \right),\label{eq:predState}
\end{align}
such that the innovation from Eq.~\eqref{eq:delayedStateMeas} can be entirely expressed in terms of either $\upd{\est{\state}}\tidx{\previdx}$ or $\pred{\est{\state}}\tidx{\curidx}$. 
The former expression uses the STM associated with a forward propagation, while the latter expression uses its inverse, corresponding to a backward propagation. 
Note that for linear continuous-time systems, the STM $\stm\tidx{\curidx,\previdx}$ is guaranteed to be invertible, and its inverse is the backward state transition matrix $\stm\tidx{\previdx,\curidx} = \stm\tidx{\curidx,\previdx}^{-1}$. 
Making these substitutions, the innovation may be expressed as either 
\begin{align}
    \innov\tidx{\curidx} 
    &= \meas\tidx{\curidx} -  \left({\linmeasmdl}\tidx{\previdx,\curidx} + {\linmeasmdl}\tidx{\curidx,\curidx} \stm\tidx{\curidx,\previdx} \right) \upd{\est{\state}}\tidx{\previdx}  - {\linmeasmdl}\tidx{\curidx,\curidx} \actionmat\tidx{\curidx} \action\tidx{\curidx},\label{eq:delayedStateInnovForward}
\end{align}
or, equivalently,
\begin{align}
    \innov\tidx{\curidx} 
    &= \meas\tidx{\curidx} -  \left({\linmeasmdl}\tidx{\previdx,\curidx}\stm\tidx{\previdx,\curidx} + {\linmeasmdl}\tidx{\curidx,\curidx} \right) \pred{\est{\state}}\tidx{\curidx}  + {\linmeasmdl}\tidx{\previdx,\curidx} \stm\tidx{\previdx,\curidx}\actionmat\tidx{\curidx} \action\tidx{\curidx}.\label{eq:delayedStateInnov}
\end{align}

The choice of Eq.~\eqref{eq:delayedStateInnovForward} or Eq.~\eqref{eq:delayedStateInnov} in the DSKF derivation leads to two different, but equivalent, formulations, and their equivalence is demonstrated in the Appendix.

Each formulation of the DSKF has its distinct tradeoffs, and the choice between them is oftentimes driven by mission constraints, compatibility with existing navigation software frameworks, and even different implementation preferences among different navigation engineers.
The forward-time DSKF avoids explicitly inverting the STM, but expresses the measurement in terms of the prior state.
In contrast, the backward-time formulation expresses the measurement solely as a function of the current predicted state, but requires evaluation of the inverse STM.
The derivation of both formulations of the DSKF update is presented in the following subsections.

\subsection{Forward-Time Formulation of the DSKF Update}\label{ssec:forwardDSKF}

The DSKF update corresponding to the formulation in Ref.~\cite{brown1997kalman} is derived using the forward-time propagation expressions.
Substituting the measurement model from Eq.~\eqref{eq:augTrueMeas} into the innovation expression from Eq.~\eqref{eq:delayedStateInnovForward} yields
\begin{align}
    \innov\tidx{\curidx} &= \begin{bmatrix} {\linmeasmdl}\tidx{\previdx,\curidx} & {\linmeasmdl}\tidx{\curidx,\curidx}\end{bmatrix}  \begin{bmatrix} {{\state}}\tidx{\previdx} \\ {{\state}}\tidx{\curidx}\end{bmatrix} + \measnoise\tidx{\curidx} -  \left({\linmeasmdl}\tidx{\previdx,\curidx} + {\linmeasmdl}\tidx{\curidx,\curidx} \stm\tidx{\curidx,\previdx} \right) \upd{\est{\state}}\tidx{\previdx}  - {\linmeasmdl}\tidx{\curidx,\curidx} \actionmat\tidx{\curidx} \action\tidx{\curidx}.
\end{align}
Next, substituting the dynamics model from Eq.~\eqref{eq:lindyn} for $\state\tidx{\curidx}$ yields an expression relating the innovation to the prior state estimate error
\begin{align}
    \innov\tidx{\curidx} &= \begin{bmatrix} {\linmeasmdl}\tidx{\previdx,\curidx} & {\linmeasmdl}\tidx{\curidx,\curidx}\end{bmatrix}  \begin{bmatrix} {{\state}}\tidx{\previdx} \\ \stm\tidx{\curidx,\previdx} \state\tidx{\previdx} + \actionmat\tidx{\curidx} \action\tidx{\curidx} + \procnoisemat\tidx{\curidx} \procnoise\tidx{\curidx} \end{bmatrix} + \measnoise\tidx{\curidx} -  \left({\linmeasmdl}\tidx{\previdx,\curidx} + {\linmeasmdl}\tidx{\curidx,\curidx} \stm\tidx{\curidx,\previdx} \right) \upd{\est{\state}}\tidx{\previdx}  - {\linmeasmdl}\tidx{\curidx,\curidx} \actionmat\tidx{\curidx} \action\tidx{\curidx},\\
    &= \left({\linmeasmdl}\tidx{\previdx,\curidx} + {\linmeasmdl}\tidx{\curidx,\curidx} \stm\tidx{\curidx,\previdx} \right) \upd{\eststateerr}\tidx{\previdx} + {\linmeasmdl}\tidx{\curidx,\curidx} \procnoisemat\tidx{\curidx} \procnoise\tidx{\curidx} + \measnoise\tidx{\curidx}  , \label{eq:forwardDSKFInnovationSimp}
\end{align}
where $\upd{\eststateerr}\tidx{\previdx}$ denotes the prior state error, which is defined as
\begin{align}
    \upd{\eststateerr}\tidx{\previdx}  &\triangleq {\state}\tidx{\previdx} - \upd{\est{\state}}\tidx{\previdx}.
\end{align}

Let us now formulate the generalized Kalman filter state update and error, thereby relating the updated state error with the predicted state error and the innovation
\begin{gather}
    \upd{\est{\state}}\tidx{\curidx} =  \pred{\est{\state}}\tidx{\curidx} + \fwdTidx{\corr{\kalmangain}}{\curidx} {\innov}\tidx{\curidx},
     \label{eq:fwdDSKFStateUpdate} \\
    \upd{\eststateerr}\tidx{\curidx} \triangleq \state\tidx{\curidx} - \upd{\est{\state}}\tidx{\curidx} = \state\tidx{\curidx} -  \pred{\est{\state}}\tidx{\curidx} - \fwdTidx{\corr{\kalmangain}}{\curidx} {\innov}\tidx{\curidx},
\end{gather}
where $\fwdTidx{\corr{\kalmangain}}{\curidx}$ denotes the Kalman gain for the forward-time DSKF, indicated by the forward arrow in the superscript.
By substituting in Eqs.~\eqref{eq:lindyn},~\eqref{eq:predStateEstimate},~ and~\eqref{eq:forwardDSKFInnovationSimp}, the updated state error can be expressed in terms of the prior state error
\begin{align}
    \upd{\eststateerr}\tidx{\curidx} &= \left(\stm\tidx{\curidx,\previdx} \state\tidx{\previdx} + \actionmat\tidx{\curidx} \action\tidx{\curidx} + \procnoisemat\tidx{\curidx} \procnoise\tidx{\curidx}\right)  -  \left( \stm\tidx{\curidx,\previdx} {\est{\state}}\tidx{\previdx} + \actionmat\tidx{\curidx} \action\tidx{\curidx} \right) - \fwdTidx{\corr{\kalmangain}}{\curidx} \left(\left({\linmeasmdl}\tidx{\previdx,\curidx} + {\linmeasmdl}\tidx{\curidx,\curidx} \stm\tidx{\curidx,\previdx} \right) \upd{\eststateerr}\tidx{\previdx} + {\linmeasmdl}\tidx{\curidx,\curidx} \procnoisemat\tidx{\curidx} \procnoise\tidx{\curidx} + \measnoise\tidx{\curidx}\right)
    , \\
    &
    = \left( \stm\tidx{\curidx,\previdx} - \fwdTidx{\corr{\kalmangain}}{\curidx} \left({\linmeasmdl}\tidx{\previdx,\curidx} + {\linmeasmdl}\tidx{\curidx,\curidx} \stm\tidx{\curidx,\previdx} \right) \right) \upd{\eststateerr}\tidx{\previdx} 
    + \left(\eye - \fwdTidx{\corr{\kalmangain}}{\curidx}{\linmeasmdl}\tidx{\curidx,\curidx} \right) \procnoisemat\tidx{\curidx} \procnoise\tidx{\curidx} 
    - \fwdTidx{\corr{\kalmangain}}{\curidx} \measnoise\tidx{\curidx}
    .
\end{align}
From here, the corresponding updated state error covariance is evaluated as
\begin{align}
    \upd{\statecov}\tidx{\curidx}
    &
    = \left( \stm\tidx{\curidx,\previdx} - \fwdTidx{\corr{\kalmangain}}{\curidx} \left({\linmeasmdl}\tidx{\previdx,\curidx} + {\linmeasmdl}\tidx{\curidx,\curidx} \stm\tidx{\curidx,\previdx} \right) \right) \upd{\statecov}\tidx{\previdx} \left( \stm\tidx{\curidx,\previdx} - \fwdTidx{\corr{\kalmangain}}{\curidx} \left({\linmeasmdl}\tidx{\previdx,\curidx} + {\linmeasmdl}\tidx{\curidx,\curidx} \stm\tidx{\curidx,\previdx} \right) \right)\trans  \nonumber 
     \\
    & \quad
    + \left(\eye - \fwdTidx{\corr{\kalmangain}}{\curidx}{\linmeasmdl}\tidx{\curidx,\curidx} \right) \procnoisecov\tidx{\curidx} \left(\eye - \fwdTidx{\corr{\kalmangain}}{\curidx}{\linmeasmdl}\tidx{\curidx,\curidx} \right)\trans
    + \fwdTidx{\corr{\kalmangain}}{\curidx} \measnoisecov\tidx{\curidx} \fwdTidx{\corr{\kalmangain}}{\curidx}\trans
    , 
\end{align}
where Eq.~\eqref{eq:procnoisecov} was applied to reintroduce the process noise covariance~$\procnoisecov\tidx{\curidx}$.
Next, rearranging terms and using Eq.~\eqref{eq:predCov_kf} to express the predicted covariance as~$\pred{\statecov}\tidx{\curidx}$ yields the following updated state error covariance expression
\begin{align}
    \upd{\statecov}\tidx{\curidx}
    &
    = 
    \pred{\statecov}\tidx{\curidx} 
    - \left(\stm\tidx{\curidx,\previdx} \upd{\statecov}\tidx{\previdx} \left({\linmeasmdl}\tidx{\previdx,\curidx} + {\linmeasmdl}\tidx{\curidx,\curidx} \stm\tidx{\curidx,\previdx} \right)\trans + \procnoisecov\tidx{\curidx}  {\linmeasmdl}\tidx{\curidx,\curidx}\trans \right) \fwdTidx{\corr{\kalmangain}}{\curidx}\trans 
    - \fwdTidx{\corr{\kalmangain}}{\curidx} \left(\left({\linmeasmdl}\tidx{\previdx,\curidx} + {\linmeasmdl}\tidx{\curidx,\curidx} \stm\tidx{\curidx,\previdx} \right) \upd{\statecov}\tidx{\previdx} \stm\tidx{\curidx,\previdx}\trans + {\linmeasmdl}\tidx{\curidx,\curidx} \procnoisecov\tidx{\curidx} \right)
     \nonumber \\ 
    & \quad
    + \fwdTidx{\corr{\kalmangain}}{\curidx} \left( {\linmeasmdl}\tidx{\curidx,\curidx} \procnoisecov\tidx{\curidx} {\linmeasmdl}\tidx{\curidx,\curidx}\trans + \left({\linmeasmdl}\tidx{\previdx,\curidx} + {\linmeasmdl}\tidx{\curidx,\curidx} \stm\tidx{\curidx,\previdx} \right) \upd{\statecov}\tidx{\previdx} \left({\linmeasmdl}\tidx{\previdx,\curidx} + {\linmeasmdl}\tidx{\curidx,\curidx} \stm\tidx{\curidx,\previdx} \right)\trans + \measnoisecov\tidx{\curidx} \right) \fwdTidx{\corr{\kalmangain}}{\curidx}\trans
    ,
\end{align}
which simplifies to
\begin{gather}
    \upd{\statecov}\tidx{\curidx} = \pred{\statecov}\tidx{\curidx} 
    - \Mcoeff\tidx{\curidx}\trans \fwdTidx{\corr{\kalmangain}}{\curidx}\trans 
    - \fwdTidx{\corr{\kalmangain}}{\curidx} \Mcoeff\tidx{\curidx}
    + \fwdTidx{\corr{\kalmangain}}{\curidx} \Lcoeff\tidx{\curidx} \fwdTidx{\corr{\kalmangain}}{\curidx}\trans 
    , \label{eq:fwdDSKFStateUpdateCov1}
\end{gather}
where
\begin{gather}
    \Lcoeff\tidx{\curidx} \triangleq {\linmeasmdl}\tidx{\curidx,\curidx} \procnoisecov\tidx{\curidx} {\linmeasmdl}\tidx{\curidx,\curidx}\trans + \left({\linmeasmdl}\tidx{\previdx,\curidx} + {\linmeasmdl}\tidx{\curidx,\curidx} \stm\tidx{\curidx,\previdx} \right) \upd{\statecov}\tidx{\previdx} \left({\linmeasmdl}\tidx{\previdx,\curidx} + {\linmeasmdl}\tidx{\curidx,\curidx} \stm\tidx{\curidx,\previdx} \right)\trans + \measnoisecov\tidx{\curidx} , \\
    \Mcoeff\tidx{\curidx} \triangleq  {\linmeasmdl}\tidx{\curidx,\curidx} \procnoisecov\tidx{\curidx} + \left({\linmeasmdl}\tidx{\previdx,\curidx} + {\linmeasmdl}\tidx{\curidx,\curidx} \stm\tidx{\curidx,\previdx} \right) \upd{\statecov}\tidx{\previdx} \stm\tidx{\curidx,\previdx}\trans .
\end{gather}
Observe here that $\Lcoeff\tidx{\curidx}$ is a symmetric matrix.
The expressions for $\Lcoeff\tidx{k}$ and $\Mcoeff\tidx{k}$ can be further simplified by substituting $\procnoisecov\tidx{\curidx} = \pred{\statecov}\tidx{\curidx} - \stm\tidx{\curidx,\previdx} \upd{\statecov}\tidx{\previdx} \stm\tidx{\curidx,\previdx}\trans$ by rearrangement of Eq.~\eqref{eq:predCov_kf}, yielding
\begin{align}
    \Lcoeff\tidx{\curidx}  
    & = {\linmeasmdl}\tidx{\curidx,\curidx} \pred{\statecov}\tidx{\curidx} {\linmeasmdl}\tidx{\curidx,\curidx}\trans
    + {\linmeasmdl}\tidx{\previdx,\curidx} \upd{\statecov}\tidx{\previdx} {\linmeasmdl}\tidx{\previdx,\curidx}\trans
    + {\linmeasmdl}\tidx{\previdx,\curidx} \upd{\statecov}\tidx{\previdx} \stm\tidx{\curidx,\previdx}\trans {\linmeasmdl}\tidx{\curidx,\curidx}\trans
    + {\linmeasmdl}\tidx{\curidx,\curidx} \stm\tidx{\curidx,\previdx} \upd{\statecov}\tidx{\previdx} {\linmeasmdl}\tidx{\previdx,\curidx}\trans
    + \measnoisecov\tidx{\curidx}
    , \label{eq:finalL}
    \\
    \Mcoeff\tidx{\curidx} 
    & = 
    {\linmeasmdl}\tidx{\curidx,\curidx} \pred{\statecov}\tidx{\curidx}
    + {\linmeasmdl}\tidx{\previdx,\curidx} \upd{\statecov}\tidx{\previdx} \stm\tidx{\curidx,\previdx}\trans \label{eq:finalM}
    .
\end{align}
Let us now solve for the optimal Kalman gain from the updated state error covariance expression in Eq.~\eqref{eq:fwdDSKFStateUpdateCov1}. 
As with the conventional Kalman filter, this is accomplished by enforcing the first-order necessary condition
\begin{gather}
    \frac{\partial \tr{\upd{\statecov}\tidx{\curidx}}}{\partial \fwdTidx{\corr{\kalmangain}}{\curidx}} = -2 \Mcoeff\tidx{\curidx}\trans + 2 \fwdTidx{\corr{\kalmangain}}{\curidx} \Lcoeff\tidx{\curidx} = \zeromat
    ,
    \\
    \fwdTidx{\corr{\kalmangain}}{\curidx} = \Mcoeff\tidx{\curidx}\trans \Lcoeff\tidx{\curidx}\inv 
    . \label{eq:fwdDSKFkalmanGain}
\end{gather} 
which has made use of the fact that $\Lcoeff\tidx{\curidx}$ is symmetric. 
Substituting this Kalman gain back into Eq.~\eqref{eq:fwdDSKFStateUpdateCov1} yields a simplified updated state error covariance
\begin{align}
    \upd{\statecov}\tidx{\curidx} &= \pred{\statecov}\tidx{\curidx} 
    - \Mcoeff\tidx{\curidx}\trans \Lcoeff\tidx{\curidx}\inv \Mcoeff\tidx{\curidx}  
    - \Mcoeff\tidx{\curidx}\trans \Lcoeff\tidx{\curidx}\inv \Mcoeff\tidx{\curidx}
    + \left( \Mcoeff\tidx{\curidx}\trans \Lcoeff\tidx{\curidx}\inv \right) \Lcoeff\tidx{\curidx} \left( \Lcoeff\tidx{\curidx}\inv \Mcoeff\tidx{\curidx}   \right) 
    , 
    \\
    &= \pred{\statecov}\tidx{\curidx} 
    - \Mcoeff\tidx{\curidx}\trans \Lcoeff\tidx{\curidx}\inv \Mcoeff\tidx{\curidx}  \label{eq:DSKF_LMupdate1}
    , 
\end{align}
Note that by alternatively substituting $\Mcoeff\tidx{\curidx} = \Lcoeff\tidx{\curidx} \fwdTidx{\corr{\kalmangain}}{\curidx}\trans$ into Eq.~\eqref{eq:fwdDSKFStateUpdateCov1} before simplifying, one obtains the equivalent covariance update expression
\begin{align}
    \upd{\statecov}\tidx{\curidx} = \pred{\statecov}\tidx{\curidx} 
    - \fwdTidx{\corr{\kalmangain}}{\curidx} \Lcoeff\tidx{\curidx} \fwdTidx{\corr{\kalmangain}}{\curidx}\trans, \label{eq:DSKF_BHupdate}
\end{align}
consistent with the form given in Ref.~\cite{brown1997kalman}.
In either case, Eq.~\eqref{eq:DSKF_LMupdate1} or Eq.~\eqref{eq:DSKF_BHupdate}, the update may be rewritten in terms of the Kalman gain
\begin{align}
    \upd{\statecov}\tidx{\curidx} &= \pred{\statecov}\tidx{\curidx} 
    - \fwdTidx{\corr{\kalmangain}}{\curidx} \Mcoeff\tidx{\curidx}  
    . \label{eq:fwdDSKFStateUpdateCov_final}
\end{align}

\subsection{Backward-Time Formulation of the DSKF Update}\label{ssec:backwardDSKF}
In some cases, the backward-time formulation is preferred for the DSKF update, as it requires only the predicted state estimate and covariance to construct the innovation and perform the update.
In this sense, it mirrors the conventional Kalman filter described in Section~\ref{sec:kf}.
This formulation corresponds to that used in Ref.~\cite{christian2025optical}.
Recall, again, that the forward-time and backward-time formulations of the DSKF are equivalent, as shown in the Appendix.

To derive the backward-time formulation of the DSKF update, let us begin 
by substituting the measurement model from Eq.~\eqref{eq:augTrueMeas} into the innovation expression from the forward-time propagation in Eq.~\eqref{eq:delayedStateInnov} to obtain
\begin{align}
    \innov\tidx{\curidx} &=  \begin{bmatrix} {\linmeasmdl}\tidx{\previdx,\curidx} & {\linmeasmdl}\tidx{\curidx,\curidx}\end{bmatrix}  \begin{bmatrix} {{\state}}\tidx{\previdx} \\ {{\state}}\tidx{\curidx}\end{bmatrix} + \measnoise\tidx{\curidx}   - \left({\linmeasmdl}\tidx{\previdx,\curidx}\stm\tidx{\previdx,\curidx} + {\linmeasmdl}\tidx{\curidx,\curidx} \right) \pred{\est{\state}}\tidx{\curidx} + {\linmeasmdl}\tidx{\previdx,\curidx} \stm\tidx{\previdx,\curidx}\actionmat\tidx{\curidx} \action\tidx{\curidx}. \label{eq:delayedStateInnov2}
\end{align}
By rearranging Eq.~\eqref{eq:lindyn}, the true prior state can be represented in terms of the the current state as
\begin{align}
    \state\tidx{\previdx} &= \stm\tidx{\previdx,\curidx} \left(\state\tidx{\curidx} - \actionmat\tidx{\curidx} \action\tidx{\curidx} - \procnoisemat\tidx{\curidx} \procnoise\tidx{\curidx}\right).
\end{align}
Substituting this expression for $\state\tidx{\previdx}$ into Eq.~\eqref{eq:delayedStateInnov2} yields the desired relationship expressing the innovation in terms of the predicted state error
\begin{gather}
    \innov\tidx{\curidx} 
    = 
    \left({\linmeasmdl}\tidx{\previdx,\curidx}\stm\tidx{\previdx,\curidx} + {\linmeasmdl}\tidx{\curidx,\curidx} \right) \pred{\eststateerr}\tidx{\curidx} 
     - {\linmeasmdl}\tidx{\previdx,\curidx}\stm\tidx{\previdx,\curidx} \procnoisemat\tidx{\curidx} \procnoise\tidx{\curidx} + \measnoise\tidx{\curidx} , \label{eq:augInnov} \\
     \pred{\eststateerr}\tidx{\curidx} \triangleq {\state}\tidx{\curidx} - \pred{\est{\state}}\tidx{\curidx}  . \label{eq:predStateErrorDef}
\end{gather}
Similar to the steps taken in Section~\ref{ssec:forwardDSKF}, let us now formulate the generalized Kalman filter state update and error as 
\begin{gather}
    \upd{\est{\state}}\tidx{\curidx} =  \pred{\est{\state}}\tidx{\curidx} + \bwdTidx{\kalmangain}{\curidx} {\innov}\tidx{\curidx},\label{eq:explStateUpdate} \\
    \upd{\eststateerr}\tidx{\curidx} \triangleq \state\tidx{\curidx} - \upd{\est{\state}}\tidx{\curidx} = \state\tidx{\curidx} -  \pred{\est{\state}}\tidx{\curidx} - \bwdTidx{\kalmangain}{\curidx} {\innov}\tidx{\curidx}, \label{eq:ecmStateUpdate}
\end{gather}
where $\bwdTidx{\kalmangain}{\curidx}$ denotes the Kalman gain for the backward-time DSKF formulation, indicated by the backward arrow in the superscript.
For simpler notation going forward, let us also define the following terms
\begin{gather}
    \corr{\linmeasmdlprev}\tidx{\curidx} \triangleq {\linmeasmdl}\tidx{\previdx,\curidx}\stm\tidx{\previdx,\curidx},
    \qquad 
    \corr{\linmeasmdl}\tidx{\curidx} \triangleq \corr{\linmeasmdlprev}\tidx{\curidx}  + {\linmeasmdl}\tidx{\curidx,\curidx}, \label{eq:corrHJDef}
    \\
    \corr{\crosscov}\tidx{\curidx} \triangleq \corr{\linmeasmdlprev}\tidx{\curidx}  \procnoisecov\tidx{\curidx},
    \qquad 
    \corr{\measnoisecov}\tidx{\curidx} \triangleq \corr{\linmeasmdlprev}\tidx{\curidx}\procnoisecov\tidx{\curidx}\corr{\linmeasmdlprev}\tidx{\curidx}\trans + \measnoisecov\tidx{\curidx}. \label{eq:corrNRDef}
\end{gather}
Substituting Eq.~\eqref{eq:augInnov} into Eq.~\eqref{eq:ecmStateUpdate} and simplifying yields
\begin{align}
    \upd{\eststateerr}\tidx{\curidx} 
    &= \left(\eye - \bwdTidx{\kalmangain}{\curidx} \left({\linmeasmdl}\tidx{\previdx,\curidx}\stm\tidx{\previdx,\curidx} + {\linmeasmdl}\tidx{\curidx,\curidx} \right) \right) \pred{\eststateerr}\tidx{\curidx} + \bwdTidx{\kalmangain}{\curidx} {\linmeasmdl}\tidx{\previdx,\curidx}\stm\tidx{\previdx,\curidx} \procnoisemat\tidx{\curidx} \procnoise\tidx{\curidx} 
    - \bwdTidx{\kalmangain}{\curidx} \measnoise\tidx{\curidx}, \label{eq:stateUpdCorr} \\
    &= \left(\eye - \bwdTidx{\kalmangain}{\curidx} \corr{\linmeasmdl}\tidx{\curidx} \right)  \pred{\eststateerr}\tidx{\curidx}
     + \bwdTidx{\kalmangain}{\curidx} \corr{\linmeasmdlprev}\tidx{\curidx} \procnoisemat\tidx{\curidx} \procnoise\tidx{\curidx} 
    - \bwdTidx{\kalmangain}{\curidx} \measnoise\tidx{\curidx}. \label{eq:stateUpdCorr2}
\end{align}
Before deriving the covariance of $\upd{\eststateerr}\tidx{\curidx}$, a relationship between the predicted state error~$\pred{\eststateerr}\tidx{\curidx}$ and the process noise~$\procnoise\tidx{\curidx}$ is first established.
Substituting Eqs.~\eqref{eq:lindyn}~and~\eqref{eq:predStateEstimate} into Eq.~\eqref{eq:predStateErrorDef} yields this relationship
\begin{align}
    \pred{\eststateerr}\tidx{\curidx} &= \stm\tidx{\curidx,\previdx}\upd{\eststateerr}\tidx{\previdx} + \procnoisemat\tidx{\curidx} \procnoise\tidx{\curidx}. \label{eq:relatePredStateProcNoise}
\end{align}
Note that the predicted state error $ \pred{\eststateerr}\tidx{\curidx}$ and process noise $\procnoise\tidx{\curidx}$ are zero-mean. 
Using Eq.~\eqref{eq:relatePredStateProcNoise}, the cross-covariance between the predicted state error and process noise is derived as
\begin{align}
    \expec{ \pred{\eststateerr}\tidx{\curidx}\procnoise\trans\tidx{\curidx} } &= \expec{\left( \stm\tidx{\curidx,\previdx}\upd{\eststateerr}\tidx{\previdx} + \procnoisemat\tidx{\curidx} \procnoise\tidx{\curidx}\right)\procnoise\trans\tidx{\curidx} }, \\
    &= \stm\tidx{\curidx,\previdx} \expec{ \upd{\eststateerr}\tidx{\previdx}\procnoise\trans\tidx{\curidx}} 
    + \procnoisemat\tidx{\curidx} \expec{ \procnoise\tidx{\curidx}\procnoise\trans\tidx{\curidx} }, \\
    &= \procnoisemat\tidx{\curidx} \unscaledprocnoisecov\tidx{\curidx},
\end{align}
where the last simplification step is due to the independence of the prior state error~$\upd{\eststateerr}\tidx{\previdx}$ from the current process noise~$\procnoise\tidx{\curidx}$. 

Let us now evaluate the updated state error covariance using Eq.~\eqref{eq:stateUpdCorr2}, noting that the measurement noise is uncorrelated with the prior process noise and state error
\begin{align}
    \upd{\statecov}\tidx{\curidx}
    &= 
    \left(\eye - \bwdTidx{\kalmangain}{\curidx} \corr{\linmeasmdl}\tidx{\curidx} \right)  \pred{\statecov}\tidx{\curidx}\left(\eye - \bwdTidx{\kalmangain}{\curidx} \corr{\linmeasmdl}\tidx{\curidx} \right) \trans + \left(\eye - \bwdTidx{\kalmangain}{\curidx} \corr{\linmeasmdl}\tidx{\curidx} \right)  \procnoisemat\tidx{\curidx} \unscaledprocnoisecov\tidx{\curidx} \left(\bwdTidx{\kalmangain}{\curidx} \corr{\linmeasmdlprev}\tidx{\curidx} \procnoisemat\tidx{\curidx}\right)\trans \nonumber
    \\
    & \quad + \left(\bwdTidx{\kalmangain}{\curidx} \corr{\linmeasmdlprev}\tidx{\curidx} \procnoisemat\tidx{\curidx}\right) \unscaledprocnoisecov\tidx{\curidx} \procnoisemat\trans\tidx{\curidx}  \left(\eye - \bwdTidx{\kalmangain}{\curidx} \corr{\linmeasmdl}\tidx{\curidx} \right)\trans 
    + \left(\bwdTidx{\kalmangain}{\curidx} \corr{\linmeasmdlprev}\tidx{\curidx} \procnoisemat\tidx{\curidx}\right)\unscaledprocnoisecov\tidx{\curidx}\left(\bwdTidx{\kalmangain}{\curidx} \corr{\linmeasmdlprev}\tidx{\curidx} \procnoisemat\tidx{\curidx}\right)\trans
    + \bwdTidx{\kalmangain}{\curidx} \measnoisecov\tidx{\curidx} \bwdTidx{\kalmangain}{\curidx}\trans.
\end{align}
Using Eq.~\eqref{eq:procnoisecov}, the process noise covariance~$\procnoisecov\tidx{k}$ is reintroduced, and the resulting expression is rearranged to recover $\corr{\crosscov}\tidx{\curidx}$ and $\corr{\measnoisecov}\tidx{\curidx}$ from~Eqs.~\eqref{eq:corrHJDef}~and~\eqref{eq:corrNRDef} 
\begin{align}
    \upd{\statecov}\tidx{\curidx} &= \left( \eye - \bwdTidx{\kalmangain}{\curidx}\corr{\linmeasmdl}\tidx{\curidx} \right) \pred{\statecov}\tidx{\curidx}\left( \eye - \bwdTidx{\kalmangain}{\curidx} \corr{\linmeasmdl}\tidx{\curidx} \right)\trans 
    + \left( \eye - \bwdTidx{\kalmangain}{\curidx} \corr{\linmeasmdl}\tidx{\curidx} \right) \corr{\crosscov}\trans\tidx{\curidx} \bwdTidx{\kalmangain}{\curidx}\trans 
    + \bwdTidx{\kalmangain}{\curidx} \corr{\crosscov}\tidx{\curidx}   \left( \eye - \bwdTidx{\kalmangain}{\curidx} \corr{\linmeasmdl}\tidx{\curidx} \right)\trans
    + \bwdTidx{\kalmangain}{\curidx} \corr{\measnoisecov}\tidx{\curidx} \bwdTidx{\kalmangain}{\curidx}\trans , \label{eq:corrStateCovUpdate}
\end{align}
where the corresponding Kalman gain can be derived as
\begin{align}
    \bwdTidx{\kalmangain}{\curidx} &= \left(\pred{\statecov}\tidx{\curidx}\corr{\linmeasmdl}\trans\tidx{\curidx} -  \corr{\crosscov}\trans\tidx{\curidx} \right)
    \left( \corr{\linmeasmdl}\tidx{\curidx}\pred{\statecov}\tidx{\curidx} \corr{\linmeasmdl}\trans\tidx{\curidx} 
    - \corr{\crosscov}\tidx{\curidx} \corr{\linmeasmdl}\trans\tidx{\curidx}
    - \corr{\linmeasmdl}\tidx{\curidx}\corr{\crosscov}\trans\tidx{\curidx}
    + \corr{\measnoisecov}\tidx{\curidx} 
    \right)\inv, \label{eq:corrKalmanGain}
\end{align}
in a similar manner as in Section~\ref{ssec:forwardDSKF}. Notice that if ${\linmeasmdl}\tidx{\previdx,\curidx}=0$ (i.e., the measurement does not depend on the prior state), then both $\corr{\linmeasmdlprev}\tidx{\curidx}$ and $\corr{\crosscov}\tidx{\curidx} \to 0$, while $\corr{\linmeasmdl}\tidx{\curidx}\to\linmeasmdl\tidx{\curidx}$ and $\corr{\measnoisecov}\tidx{\curidx}\to\measnoisecov\tidx{\curidx}$, recovering the traditional Kalman filter update in Eq.~\eqref{eq:corrStateCovUpdate} and classical Kalman gain in Eq.~\eqref{eq:corrKalmanGain}.

%% file: sections/7_equivalence.tex
\section{Equivalence of Stochastic Cloning and the Delayed-State Kalman Filter}\label{sec:equivalence}

This section proves the equivalence of stochastic cloning and the delayed-state Kalman filter by induction. 
The analysis is carried out with respect to the forward-time formulation of the DSKF from Section~\ref{ssec:forwardDSKF}, while the equivalence between the two DSKF formulations is demonstrated in the Appendix.
The base case of the proof by induction is trivial, as both filters start with the same belief at~$\timeidx{0}$. 
For the inductive step, assuming equivalence at time~$\timeidx{\previdx}$, equivalence is shown to be maintained at the subsequent update time~$\timeidx{\curidx}$.
Section~\ref{ssec:equivStateUpdate} shows equivalence of the updated state estimate, and Section~\ref{ssec:equivCovUpdate} demonstrates the equivalence of the covariance update.

\subsection{Equivalence of the State Update} \label{ssec:equivStateUpdate}
Let us examine the state update for both filters in Eqs.~\eqref{eq:scStateUpdate}~and~\eqref{eq:fwdDSKFStateUpdate}. 
Given the prior updated estimate $\upd{\est{\state}}\tidx{\previdx}$, both delayed-state filters yield the same predicted state $\pred{\est{\state}}\tidx{\curidx}$, since the prediction step follows the same dynamics.
The innovation ${\innov}\tidx{\curidx}$ is also identical, so verifying the equivalence of the Kalman gain confirms the equivalence of state updates.

First, consider the stochastic cloning Kalman gain $\aug{\kalmangain}\tidx{\curidx,\curidx}$, which from Eq.~\eqref{eq:scAugKalmanGain} is the bottom submatrix of the augmented Kalman gain $\aug{\kalmangain}\tidx{\curidx}$.
From Eq.~\eqref{eq:sckalmangain}, the augmented matrix definitions are substituted before extracting the bottom submatrix $\aug{\kalmangain}\tidx{\curidx,\curidx}$
\begin{align}
    \aug{\kalmangain}\tidx{\curidx} &= \begin{bmatrix}
        \cdot & \cdot  \\
        \stm\tidx{\curidx,\previdx} \upd{\statecov}\tidx{\previdx} & \pred{\statecov}\tidx{\curidx}
     \end{bmatrix} 
     \begin{bmatrix} {\linmeasmdl}\trans\tidx{\previdx,\curidx} \\ {\linmeasmdl}\trans\tidx{\curidx,\curidx} \end{bmatrix} \left( 
     \begin{bmatrix} {\linmeasmdl}\tidx{\previdx,\curidx} & {\linmeasmdl}\tidx{\curidx,\curidx} \end{bmatrix} 
     \begin{bmatrix}
        \upd{\statecov}\tidx{\previdx} & \upd{\statecov}\tidx{\previdx} \stm\trans\tidx{\curidx,\previdx}  \\
        \stm\tidx{\curidx,\previdx} \upd{\statecov}\tidx{\previdx} & \pred{\statecov}\tidx{\curidx}
     \end{bmatrix} 
     \begin{bmatrix} {\linmeasmdl}\trans\tidx{\previdx,\curidx} \\ {\linmeasmdl}\trans\tidx{\curidx,\curidx} \end{bmatrix}
     + \measnoisecov\tidx{\curidx}
     \right)\inv. \label{eq:equivKalmanGain}
\end{align}
Writing out the the matrix expressions for the stochastic cloning Kalman gain submatrix $\aug{\kalmangain}\tidx{\curidx,\curidx}$ demonstrates its equivalence with the forward-time DSKF Kalman gain
\begin{align}
    \aug{\kalmangain}\tidx{\curidx,\curidx} 
    & = \left(\pred{\statecov}\tidx{\curidx} {\linmeasmdl}\trans\tidx{\curidx,\curidx} + \stm\tidx{\curidx,\previdx} \upd{\statecov}\tidx{\previdx} {\linmeasmdl}\trans\tidx{\previdx,\curidx} \right) 
    \left( 
    \begin{bmatrix} \left({\linmeasmdl}\tidx{\previdx,\curidx} \upd{\statecov}\tidx{\previdx} + {\linmeasmdl}\tidx{\curidx,\curidx} \stm\tidx{\curidx,\previdx} \upd{\statecov}\tidx{\previdx}\right)  & 
    \left(  {\linmeasmdl}\tidx{\previdx,\curidx} \upd{\statecov}\tidx{\previdx} \stm\trans\tidx{\curidx,\previdx}
    + {\linmeasmdl}\tidx{\curidx,\curidx} \pred{\statecov}\tidx{\curidx} \right) \end{bmatrix} 
    \begin{bmatrix} {\linmeasmdl}\trans\tidx{\previdx,\curidx} \\ {\linmeasmdl}\trans\tidx{\curidx,\curidx} \end{bmatrix} 
    + \measnoisecov\tidx{\curidx} 
    \right)\inv
    , 
    \\
    & = 
    \left( {\linmeasmdl}\tidx{\curidx,\curidx}\pred{\statecov}\tidx{\curidx} +  {\linmeasmdl}\tidx{\previdx,\curidx} \upd{\statecov}\tidx{\previdx}  \stm\tidx{\curidx,\previdx}\trans \right)\trans 
    \left(
    {\linmeasmdl}\tidx{\previdx,\curidx} \upd{\statecov}\tidx{\previdx} {\linmeasmdl}\trans\tidx{\previdx,\curidx} 
    + {\linmeasmdl}\tidx{\curidx,\curidx} \stm\tidx{\curidx,\previdx} \upd{\statecov}\tidx{\previdx} {\linmeasmdl}\trans\tidx{\previdx,\curidx}
    +  {\linmeasmdl}\tidx{\previdx,\curidx} \upd{\statecov}\tidx{\previdx} \stm\trans\tidx{\curidx,\previdx} {\linmeasmdl}\trans\tidx{\curidx,\curidx}
    + {\linmeasmdl}\tidx{\curidx,\curidx} \pred{\statecov}\tidx{\curidx} {\linmeasmdl}\trans\tidx{\curidx,\curidx}
    + \measnoisecov\tidx{\curidx} 
    \right)\inv
    ,
    \\
    & =
    \Mcoeff\tidx{\curidx}\trans \Lcoeff\tidx{\curidx}\inv
    ,
    \\
    & = \fwdTidx{\corr{\kalmangain}}{\curidx},  \label{eq:equivalentKalmanGain}
\end{align} 
where the substitution of $\Lcoeff\tidx{\curidx}$~and~$\Mcoeff\tidx{\curidx}$ uses Eqs.~\eqref{eq:finalL}~and~\eqref{eq:finalM}, respectively. 

\subsection{Equivalence of the Updated State Error Covariance}\label{ssec:equivCovUpdate}
To compare the updated state error covariance between the two delayed-state filters, let us start with the stochastic cloning filter covariance in Eq.~\eqref{eq:stateUpdateSC}, which corresponds to the bottom right $\numstates \times \numstates$ submatrix of the augmented covariance $\aug{\statecov}\tidx{\curidx}$, i.e.,
\begin{align}
    \upd{\aug{\statecov}}\tidx{\curidx} = \begin{bmatrix} \cdot & \cdot \\ \cdot & \upd{\aug{\statecov}}\tidx{\curidx,\curidx} \end{bmatrix}.
\end{align}
To extract the corresponding submatrix $\upd{\aug{\statecov}}\tidx{\curidx,\curidx}$, Eq.~\eqref{eq:scCovFinal} is expanded with the definitions for $\Smat$, $\aug{\linmeasmdl}\tidx{\curidx}$, and $\pred{\aug{\statecov}}\tidx{\curidx}$ before multiplying out the relevant submatrix terms
\begin{align}
    \upd{\aug{\statecov}}\tidx{\curidx,\curidx} 
    &= \left( \Smat - \aug{\kalmangain}\tidx{\curidx,\curidx} \aug{\linmeasmdl}\tidx{\curidx} \right) \pred{\aug{\statecov}}\tidx{\curidx}\left( \Smat - \aug{\kalmangain}\tidx{\curidx,\curidx} \aug{\linmeasmdl}\tidx{\curidx} \right)\trans +  \aug{\kalmangain}\tidx{\curidx,\curidx} \measnoisecov\tidx{\curidx} \aug{\kalmangain}\tidx{\curidx,\curidx}\trans
    ,
    \\
    & =
    \aug{\kalmangain}\tidx{\curidx,\curidx} {\linmeasmdl}\tidx{\previdx,\curidx} \upd{\statecov}\tidx{\previdx} {\linmeasmdl}\tidx{\previdx,\curidx}\trans \aug{\kalmangain}\tidx{\curidx,\curidx}\trans
    - \left(\eyedim{\numstates} - \aug{\kalmangain}\tidx{\curidx,\curidx} {\linmeasmdl}\tidx{\curidx,\curidx}\right) \stm\tidx{\curidx,\previdx} \upd{\statecov}\tidx{\previdx} {\linmeasmdl}\tidx{\previdx,\curidx}\trans \aug{\kalmangain}\tidx{\curidx,\curidx}\trans
    - \aug{\kalmangain}\tidx{\curidx,\curidx} {\linmeasmdl}\tidx{\previdx,\curidx} \upd{\statecov}\tidx{\previdx} \stm\tidx{\curidx,\previdx}\trans \left(\eyedim{\numstates} - \aug{\kalmangain}\tidx{\curidx,\curidx} {\linmeasmdl}\tidx{\curidx,\curidx}\right)\trans
     \nonumber \\
    & \quad 
    + \left(\eyedim{\numstates} - \aug{\kalmangain}\tidx{\curidx,\curidx} {\linmeasmdl}\tidx{\curidx,\curidx}\right) \pred{\statecov}\tidx{\curidx} 
    \left(\eyedim{\numstates} - \aug{\kalmangain}\tidx{\curidx,\curidx} {\linmeasmdl}\tidx{\curidx,\curidx}\right)\trans
    +  \aug{\kalmangain}\tidx{\curidx,\curidx} \measnoisecov\tidx{\curidx} \aug{\kalmangain}\tidx{\curidx,\curidx}\trans
    .
\end{align}
where further expansion and rearrangement reveal terms corresponding to $\Lcoeff\tidx{\curidx}$ and $\Mcoeff\tidx{\curidx}$, as defined in Eqs.~\eqref{eq:finalL}~and~\eqref{eq:finalM}
\begin{align}
    \upd{\aug{\statecov}}\tidx{\curidx,\curidx} 
    &
    =
    \pred{\statecov}\tidx{\curidx} 
    - 
    \aug{\kalmangain}\tidx{\curidx,\curidx} \left( 
    {\linmeasmdl}\tidx{\curidx,\curidx} \pred{\statecov}\tidx{\curidx} 
    + {\linmeasmdl}\tidx{\previdx,\curidx} \upd{\statecov}\tidx{\previdx} \stm\tidx{\curidx,\previdx}\trans
    \right)
    -
    \left( 
     \pred{\statecov}\tidx{\curidx} {\linmeasmdl}\tidx{\curidx,\curidx}\trans 
    + \stm\tidx{\curidx,\previdx} \upd{\statecov}\tidx{\previdx}  {\linmeasmdl}\tidx{\previdx,\curidx}\trans
    \right)
    \aug{\kalmangain}\tidx{\curidx,\curidx}\trans
     \nonumber \\
    & \quad 
    + 
    \aug{\kalmangain}\tidx{\curidx,\curidx} \left( 
    {\linmeasmdl}\tidx{\curidx,\curidx} \pred{\statecov}\tidx{\curidx} {\linmeasmdl}\tidx{\curidx,\curidx}\trans 
    + {\linmeasmdl}\tidx{\previdx,\curidx} \upd{\statecov}\tidx{\previdx}  {\linmeasmdl}\tidx{\previdx,\curidx}\trans
    + {\linmeasmdl}\tidx{\curidx,\curidx} \stm\tidx{\curidx,\previdx} \upd{\statecov}\tidx{\previdx}  {\linmeasmdl}\tidx{\previdx,\curidx}\trans 
    + {\linmeasmdl}\tidx{\previdx,\curidx} \upd{\statecov}\tidx{\previdx} \stm\tidx{\curidx,\previdx}\trans {\linmeasmdl}\tidx{\curidx,\curidx}\trans 
    + \measnoisecov\tidx{\curidx} \right) \aug{\kalmangain}\tidx{\curidx,\curidx}\trans
    ,
    \\
    &
    =
    \pred{\statecov}\tidx{\curidx} 
    -
    \aug{\kalmangain}\tidx{\curidx,\curidx} \Mcoeff\tidx{\curidx}
    - \Mcoeff\tidx{\curidx}\trans \aug{\kalmangain}\tidx{\curidx,\curidx}\trans
    + \aug{\kalmangain}\tidx{\curidx,\curidx} \Lcoeff\tidx{\curidx} \aug{\kalmangain}\tidx{\curidx,\curidx}\trans
    .
\end{align}
Recall from Eq.~\eqref{eq:equivalentKalmanGain} that the stochastic cloning and DSKF Kalman gains are equivalent.
Leveraging this and Eq.~\eqref{eq:fwdDSKFkalmanGain} yields the final reformulated expression of the updated state error covariance
\begin{align}
    \upd{\aug{\statecov}}\tidx{\curidx,\curidx} 
    &
    =
    \pred{\statecov}\tidx{\curidx} 
    -
    \Mcoeff\tidx{\curidx}\trans \Lcoeff\tidx{\curidx}\inv \Mcoeff\tidx{\curidx}
    - \Mcoeff\tidx{\curidx}\trans  \Lcoeff\tidx{\curidx}\inv \Mcoeff\tidx{\curidx}
    + \left( \Mcoeff\tidx{\curidx}\trans \Lcoeff\tidx{\curidx}\inv \right) \Lcoeff\tidx{\curidx} \left( \Lcoeff\tidx{\curidx}\inv \Mcoeff\tidx{\curidx} \right)
    ,
    \\
    &
    =
    \pred{\statecov}\tidx{\curidx} 
    -
    \Mcoeff\tidx{\curidx}\trans \Lcoeff\tidx{\curidx}\inv \Mcoeff\tidx{\curidx}
    ,
    \\
    &
    =
    \pred{\statecov}\tidx{\curidx} 
    -
    \fwdTidx{\kalmangain}{\curidx} \Mcoeff\tidx{\curidx}
    ,
\end{align}
which is equivalent to the updated state error covariance in the DSKF in Eq.~\eqref{eq:fwdDSKFStateUpdateCov_final}.
This establishes the equivalence of the state and covariance update, thereby completing the proof of the inductive step.

The base case of the proof by induction also holds by definition, as the state and covariance at time~$\timeidx{0}$ are initialized identically, thereby completing the inductive proof of the equivalence of the two filters. 
While presented for the linear Kalman filter, the proof also holds for the extended Kalman filter via similar steps, after linearizing about the current state estimate.

%% file: sections/8_fewerClonedStates.tex
\section{Simplifications Exploiting Sparse Prior-State Dependence} \label{sec:simplifications}
In some cases, the delayed-state measurement does not depend on all prior states, but rather a subset of prior states.
For example, a delayed-state measurement which observes a relative change in position only depends on three prior states (corresponding to the prior three-dimensional position vector), whereas the $\numstates$-dimensional state vector oftentimes comprises additional states, such as velocity, attitude, bias states, and so forth.

Consider $\numprior$ as the dimension of the prior state dependency (i.e., the number of prior states on which the measurement depends), where $1 \leq \numprior \leq \numstates$.
Without loss of generality, let us assume that the first $\numprior$ states correspond to the relevant subset of the prior states.
Thus, the measurement model of Eq.~\eqref{eq:augTrueMeas} can be rewritten as
\begin{align}
    \meas\tidx{\curidx} &=  \begin{bmatrix} \ellMeasMdl & {\linmeasmdl}\tidx{\curidx,\curidx}\end{bmatrix}  \begin{bmatrix} \ellPriorState \\ {{\state}}\tidx{\curidx}\end{bmatrix} + \measnoise\tidx{\curidx}, \label{eq:augTrueMeas_ell} 
\end{align}
where $\ellPriorState$ corresponds to the relevant $\numprior$~elements of the state and $\ellMeasMdl$ corresponds to the relevant sumbatrix of the prior measurement covariance, both of which can be defined with existing terms and a projection matrix
\begin{align}
    \ellProjMat &\triangleq \begin{bmatrix} \eyedim{\numprior} & \zeromatdim{\numprior \times (\numstates - \numprior)}\end{bmatrix},
\end{align}
where $\ellPriorState \triangleq \ellProjMat \state\tidx{\previdx}$ and $\ellMeasMdl \triangleq {\linmeasmdl}\tidx{\previdx,\curidx} \ellProjMat\trans$.
Note that these definitions of $\ellPriorState$ and $\ellMeasMdl$ correspond to extracting the relevant rows and columns, respectively, and do not introduce any additional computational cost in implementation.
Indeed, the expressions from Sections~\ref{sec:sc}~and~\ref{sec:brownAndHwangFilter} are always valid and can still be used in cases where $\numprior \leq \numstates$.
In particular, in this setting, the measurement matrix coefficient with respect to the prior state $\linmeasmdl\tidx{\previdx,\curidx}$ (or the augmented measurement matrix $\aug{\linmeasmdl}\tidx{\curidx}$ in SC) becomes sparse with multiple all-zero rows, and the full prior state is still cloned in SC.
However, further simplifications can be performed in both SC and the DSKF formulations to reduce computational cost and memory.
This section examines these simplifications for each of the aforementioned filter formulations.

\subsection{Simplifications to Stochastic Cloning with Reduced Prior-State Dependence} \label{ssec:ellSC}

Given that the measurement only depends on $\numprior$~prior states, the augmented state from Eq.~\eqref{eq:augStateDef} reduces to
\begin{gather}
    \ellaugPredState = \begin{bmatrix} \ellPriorStateEst \\ \pred{\est{\state}}\tidx{\curidx} \end{bmatrix} \in \Rdim{\numprior + \numstates}. 
\end{gather} 
Correspondingly, the augmented covariance from Eq.~\eqref{eq:augCovDef} reduces to
\begin{gather}
    \ellaugPredStateCov = \begin{bmatrix}
        \ellPriorStateCov & \left(\ellPriorStateCovCol\right)\trans \stm\trans\tidx{\curidx,\previdx}  \\
        \stm\tidx{\curidx,\previdx} \ellPriorStateCovCol & \pred{\statecov}\tidx{\curidx} 
     \end{bmatrix} \in \Rdim{(\numprior + \numstates) \times (\numprior + \numstates)}, \label{eq:ellSCAugP}
\end{gather}
where $ \ellPriorStateCov \triangleq \ellProjMat \upd{\statecov}\tidx{\previdx} \ellProjMat\trans \in \Rdim{\numprior \times \numprior}$, which corresponds to the estimate error covariance of~$\ellPriorState$,~and~$
 \ellPriorStateCovCol \triangleq \ellProjMat \upd{\statecov}\tidx{\previdx} \in \Rdim{\numstates \times \numprior}$, which corresponds to the cross-covariance between $\upd{\est{\state}}\tidx{\previdx}$ and $\ellPriorState$. 
Note that these submatrix definitions involving the projection matrix~$\ellProjMat$ correspond to extracting the relevant rows and columns of $\upd{\statecov}\tidx{\previdx}$ and do not increase the computational complexity of the filtering method.

From the simplified measurement model in Eq.~\eqref{eq:augTrueMeas_ell}, let us denote the corresponding reduced-dimension augmented measurement matrix as
\begin{align}
    \ellAugH &\triangleq \begin{bmatrix} \ellMeasMdl & {\linmeasmdl}\tidx{\curidx,\curidx}\end{bmatrix} \in \Rdim{\nummeas \times \numprior}.
\end{align}
With this reduced measurement matrix coefficient expression, the innovation covariance, Kalman gain, and updated state error covariance expressions from Eqs.~\eqref{eq:scInnovCov},~\eqref{eq:sckalmangain}~and~\eqref{eq:scCovFinal} can also be simplified by operating only on the relevant submatrices
\begin{align}
    {\innovcov}\tidx{\curidx} 
    &= \ellAugH \ellaugPredStateCov \left(\ellAugH\right)\trans + \measnoisecov\tidx{\curidx}, \label{eq:ellSCInnovCov} \\
    \aug{\kalmangain}\tidx{\curidx,\curidx} 
    &= \left(\ellSmat \ellaugPredStateCov\right) \left(\ellAugH\right)\trans {\innovcov}\tidx{\curidx}\inv , \label{eq:ellSCKalman1} \\
    &= \begin{bmatrix} \stm\tidx{\curidx,\previdx} \ellPriorStateCovCol & \pred{\statecov}\tidx{\curidx} \end{bmatrix} \left(\ellAugH\right)\trans {\innovcov}\tidx{\curidx}\inv, \label{eq:ellSCKalman2}
    \\
    \upd{\aug{\statecov}}\tidx{\curidx,\curidx} &= \left( \ellSmat - \aug{\kalmangain}\tidx{\curidx,\curidx} \ellAugH \right)  \ellaugPredStateCov  \left( \ellSmat - \aug{\kalmangain}\tidx{\curidx,\curidx} \ellAugH \right)\trans +  \aug{\kalmangain}\tidx{\curidx,\curidx} \measnoisecov\tidx{\curidx} \aug{\kalmangain}\tidx{\curidx,\curidx}\trans, \label{eq:ellSCUpdCov}
\end{align}
where $\ellSmat \triangleq \begin{bmatrix} \zeromatdim{\numstates \times \numprior} & \eyedim{\numstates} \end{bmatrix}$ and where the multiplicative term $\left(\ellSmat \ellaugPredStateCov\right)$ in Eq.~\eqref{eq:ellSCKalman1} corresponds to extracting the $\numstates$~lower rows of~$\ellaugPredStateCov$, as also expressed in Eqs.~\eqref{eq:ellSCKalman1}-\eqref{eq:ellSCKalman2}.
Note that as $\numprior$ decreases, the corresponding reduced dimension of several intermediate matrices in Eqs.~\eqref{eq:ellSCInnovCov}-\eqref{eq:ellSCUpdCov} leads to lower computational cost and memory requirements for the update step.
The scaling of these costs with respect to the state, prior-state measurement dependency, and measurement dimensions (i.e., $\numstates$,~$\numprior$,~and~$\nummeas$) are further explored in Section~\ref{sec:compAndMemory}.

\subsection{Simplifications to the Forward-Time Delayed-State Kalman Filter with Reduced Prior-State Dependence}

In the case that the measurement model only depends on $\numprior$~prior states, as in Eq.~\eqref{eq:augTrueMeas_ell}, the last $\numstates - \numprior$ rows of the prior state measurement matrix ${\linmeasmdl}\tidx{\previdx,\curidx}$ become identically zero. 
Correspondingly, the expressions for
$\Lcoeff\tidx{\curidx}$~and~$\Mcoeff\tidx{\curidx}$ in Eqs.~\eqref{eq:finalL}~and~\eqref{eq:finalM} simplify by only operating on the relevant submatrices
\begin{align}
    \Lcoeff\tidx{\curidx}  
    & = {\linmeasmdl}\tidx{\curidx,\curidx} \pred{\statecov}\tidx{\curidx} {\linmeasmdl}\tidx{\curidx,\curidx}\trans
    + \ellMeasMdl \ellPriorStateCov \left( \ellMeasMdl \right)\trans
    + \ellMeasMdl \left(\ellPriorStateCovCol\right)\trans \stm\tidx{\curidx,\previdx}\trans {\linmeasmdl}\tidx{\curidx,\curidx}\trans
    + {\linmeasmdl}\tidx{\curidx,\curidx} \stm\tidx{\curidx,\previdx} \ellPriorStateCovCol \left(\ellMeasMdl\right)\trans
    + \measnoisecov\tidx{\curidx}
    ,
    \\
    \Mcoeff\tidx{\curidx} 
    & = 
    {\linmeasmdl}\tidx{\curidx,\curidx} \pred{\statecov}\tidx{\curidx}
    + \ellMeasMdl \left(\ellPriorStateCovCol\right)\trans \stm\tidx{\curidx,\previdx}\trans 
    ,
\end{align}
where $\ellPriorStateCov$~and~$\ellPriorStateCovCol$ are defined in Section~\ref{ssec:ellSC}.
These simplified definitions of $\Lcoeff\tidx{\curidx}$~and~$\Mcoeff\tidx{\curidx}$ are subsequently used to compute the Kalman gain, updated state, and updated state error covariance using the same expressions as in Eqs.~\eqref{eq:fwdDSKFkalmanGain},~\eqref{eq:fwdDSKFStateUpdate},~and~\eqref{eq:fwdDSKFStateUpdateCov_final}, respectively.

\subsection{Simplifications to the Backward-Time Delayed-State Kalman Filter with Reduced Prior-State Dependence}

For the backward-time DSKF, the evaluation of the intermediate matrix $\corr{\linmeasmdlprev}\tidx{\curidx}$ in Eq.~\eqref{eq:corrHJDef} can be simplified to
\begin{align}
    \corr{\linmeasmdlprev}\tidx{\curidx} = \ellMeasMdl \ellSTMInv,
\end{align}
where $\ellSTMInv \triangleq \ellProjMat \stm\tidx{\previdx,\curidx} \in \Rdim{\numprior \times \numstates}$ corresponds to the first $\numprior$~rows of the inverse state transition matrix.
After computing~$\corr{\linmeasmdlprev}\tidx{\curidx}$, the remaining intermediate matrices $\corr{\linmeasmdl}\tidx{\curidx}$, $\corr{\crosscov}\tidx{\curidx}$, and $\corr{\measnoisecov}\tidx{\curidx}$ are obtained using the same expressions as in Eqs.~\eqref{eq:corrHJDef}~and~\eqref{eq:corrNRDef}.
The Kalman gain, updated state, and updated state error covariance then follow from the same expressions as in Eqs.~\eqref{eq:corrKalmanGain},~\eqref{eq:explStateUpdate},~and~\eqref{eq:corrStateCovUpdate}, respectively.

%% file: sections/9_computationsMemory.tex
\section{Computational and Memory Considerations}\label{sec:compAndMemory}

To support a comparison between stochastic cloning and the delayed-state Kalman filter, arithmetic complexity is estimated using floating-point operation (flop) counts and memory usage is estimated by counting the maximum storage required for every matrix computation in each algorithm. These metrics provide general architecture-independent comparisons of algorithmic structure, whereas actual runtime performance may depend on memory hierarchy and hardware characteristics.

Table~\ref{tab1} summarizes the arithmetic complexity of each algorithm in terms of flop counts. An addition, a multiplication, and a division are each counted as one flop. Subtraction and addition operations are grouped under the number of additions, while multiplication and division operations are grouped under the number of multiplications.  These counts are based on an efficient computation sequence for the Kalman gain, state update, and covariance update. For example, in the stochastic cloning covariance update of Eq.~\eqref{eq:scCovFinal}, the flops for computing the quantity $\left( \Smat - \aug{\kalmangain}\tidx{\curidx,\curidx} \aug{\linmeasmdl}\tidx{\curidx} \right)$ are counted only once, assuming the result is reused for the transpose. These estimates further assume LU decomposition for matrix inversion~\cite{laug,fortran}, where the computational complexity of a matrix inversion consists of approximately $N^3$ adds and $N^3$ multiplies~\cite{fortran}, where $N$ is the dimension of the square matrix.

Memory utilization was estimated by counting the number of floating point values needed to store initial, intermediate, and final matrix quantities throughout the computation sequence. It is assumed that element-wise operations (e.g., matrix addition/subtraction) are performed in-place and therefore do not incur additional memory costs beyond the operands.  While compilers may further eliminate or reuse intermediate storage, this dimension-based accounting provides a conservative upper bound on peak memory usage. For matrix inversion, the use of the LAPACK function \texttt{dgetri} is assumed for an in-place inversion based on the LU decomposition, thus the memory overhead is assumed to be the lower bound of $2n$~\cite{laug}. 
The memory allocation results for stochastic cloning and both formulations of the delayed-state Kalman filter are shown in Table~\ref{tab:memtable}.

The arithmetic complexity for each algorithm is summarized as follows from Table~\ref{tab1}. For a system with $n$~states, $\ell$~cloned states, and $m$~delayed-state measurements,   each algorithm exhibits cubic scaling in the variables, i.e., $\mathcal{O}(N^3)$ where $N=n+\ell+m$. Specifically, SC has arithmetic complexity
\begin{align}
    \mathcal{O}(\ell^2m + \ell^2n + \ell m^2 + \ell mn + \ell n^2 + m^3 + m^2n + mn^2 + n^3), 
\end{align}
the forward-time DSKF has arithmetic complexity
\begin{align}
    \mathcal{O}(\ell m^2 + \ell mn + m^3 + m^2n +mn^2), 
\end{align}
and the backward-time DSKF has arithmetic complexity
\begin{align}
    \mathcal{O}(\ell mn + m^3 + m^2n + mn^2 + n^3).
\end{align}

Memory requirements for each algorithm are summarized as follows from Table~\ref{tab:memtable}. In number of floating point values, each algorithm exhibits quadratic scaling in the variables, i.e., $\mathcal{O}(N^2)$. Specifically, SC has memory complexity
\begin{align}
    \mathcal{O}(\ell m + \ell n + m^2 + mn + n^2),
\end{align} 
the forward-time DSKF has memory complexity
\begin{align}
    \mathcal{O}(\ell m + \ell n + m^2 + mn + n^2),
\end{align} 
and the backward-time DSKF has memory complexity
\begin{align}
    \mathcal{O}(\ell m + m^2 + mn + n^2 ).
\end{align} 

\begin{figure}
\centering
\includegraphics[width=\textwidth]{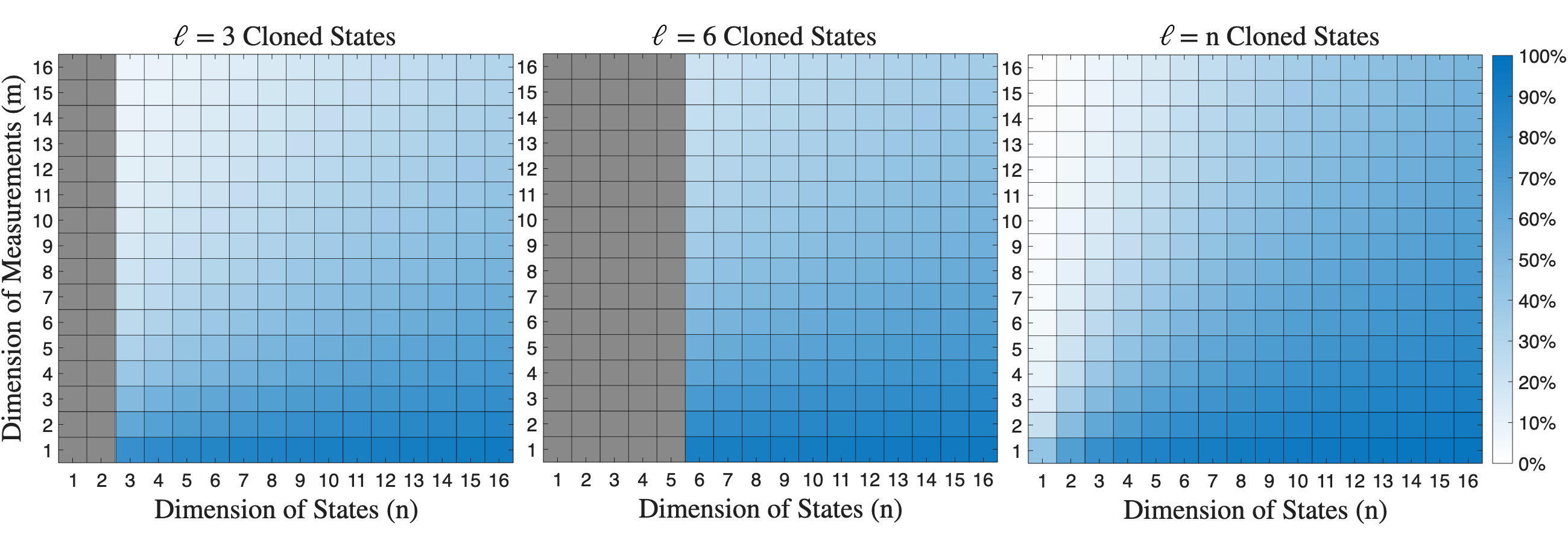}
\caption{Percent reduction in arithmetic complexity of the forward-time DSKF compared to SC for $\ell=3$, $\ell=6$, and $\ell=n$ cloned states, from left to right. DSKF consistently requires fewer flops per iteration, for all $n$ and $m$.}
\label{fig1}
\end{figure}

\begin{figure}
\centering
\includegraphics[width=\textwidth]{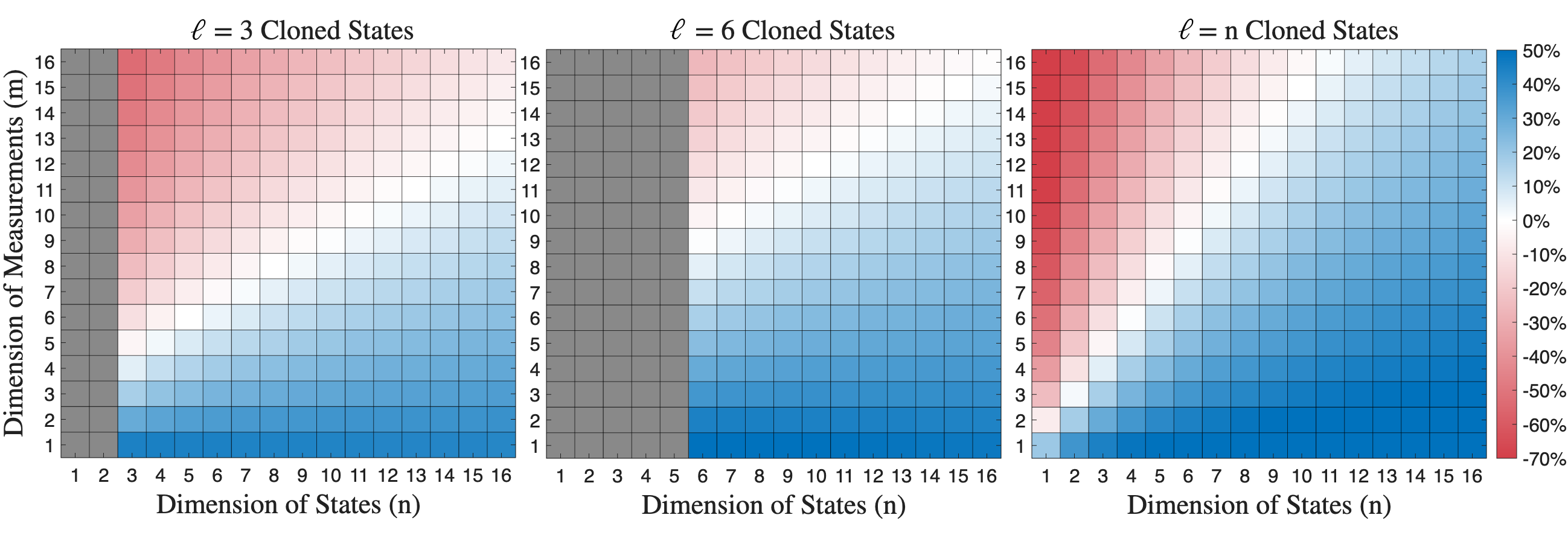}
\caption{Percent reduction in memory allocation of the forward-time DSKF in comparison to SC for $\ell=3$, $\ell=6$, and $\ell=n$ cloned states, from left to right. Blue indicates cases where the DSKF has comparatively lower memory requirements in comparison to SC, while red indicates cases where it has higher memory requirements.}
\label{fig2}
\end{figure}

Thus, it is demonstrated that SC and both formulations of the DSKF are similar algorithms in terms of algorithmic and memory complexity. For a visual comparison, Figs.~\ref{fig1}~and~\ref{fig2} illustrate a heat map of the comparative arithmetic complexity and memory allocation of the forward-time DSKF with respect to SC for various values of $n$ and $m$ and fixed values of $\ell$ cloned states. Note that relative efficiency of these algorithms is specific to the details of their implementation. 
Nevertheless, in Fig.~\ref{fig2}, a clear trend emerges when memory requirements become advantageous for the forward-time DSKF formulation over SC, which occurs for larger $n$ in comparison to $m$.
At the same time, as visualized in Fig.~\ref{fig1}, the arithmetic complexity of the forward-time DSKF is consistently lower than that of SC.

\newcommand{\bDSKFkOneM}{\ell mn + m^3 + 4m^2n + 2mn^2}
\newcommand{\bDSKFkTwoM}{ + n^3}
\newcommand{\bDSKFkThreeM}{}

\newcommand{\fDSKFkOneM}{\ell m^2 + \ell mn + m^3 + 3m^2n}
\newcommand{\fDSKFkTwoM}{ + 2mn^2 }
\newcommand{\fDSKFkThreeM}{}

\newcommand{\bDSKFxOneM}{\ell m + 2mn}

\newcommand{\fDSKFxOneM}{\ell m + 2mn}

\newcommand{\bDSKFpOneM}{m^2n + 3mn^2 + 3n^3}
\newcommand{\bDSKFpTwoM}{}
\newcommand{\bDSKFpThreeM}{}

\newcommand{\fDSKFpOneM}{mn^2}
\newcommand{\fDSKFpTwoM}{}
\newcommand{\fDSKFpThreeM}{}

\newcommand{\bDSKFkOneA}{\ell mn + m^3 + 4m^2n + m^2}
\newcommand{\bDSKFkTwoA}{+ 2mn^2 - 2mn + n^3}
\newcommand{\bDSKFkThreeA}{}

\newcommand{\fDSKFkOneA}{\ell m^2 + \ell mn + m^3 + 3m^2n}
\newcommand{\fDSKFkTwoA}{ + 2mn^2 - 3mn}
\newcommand{\fDSKFkThreeA}{}

\newcommand{\bDSKFxOneA}{\ell m + 2mn}

\newcommand{\fDSKFxOneA}{\ell m + 2mn}

\newcommand{\bDSKFpOneA}{m^2n + 3mn^2 - mn + 3n^3}
\newcommand{\bDSKFpTwoA}{- 3n^2 + n}
\newcommand{\bDSKFpThreeA}{}

\newcommand{\fDSKFpOneA}{mn^2}
\newcommand{\fDSKFpTwoA}{}
\newcommand{\fDSKFpThreeA}{}

\newcommand{\bDSKFTotOne}{2 \ell mn + 2 \ell m + 2m^3 + 10m^2n}
\newcommand{\bDSKFTotTwo}{+ m^2 + 10mn^2 + mn + 8n^3}
\newcommand{\bDSKFTotThree}{- 3n^2 + n}
\newcommand{\bDSKFTotFour}{}

\newcommand{\fDSKFTotOne}{2 \ell m^2 + 2 \ell mn + 2 \ell m + 2m^3}
\newcommand{\fDSKFTotTwo}{ + 6m^2n + 6mn^2 + mn}
\newcommand{\fDSKFTotThree}{}
\newcommand{\fDSKFTotFour}{}

\def\arraystretch{1.3}
\begin{table}
\centering
\caption{Arithmetic Complexity of Stochastic Cloning and Delayed-State Kalman Filter for $n$~States, $\ell$~Cloned States, and $m$~Measurements. The number of multiplication and addition operations are shown separately, then the total floating point operators (flops).}
\setlength{\tabcolsep}{20pt}
\begin{tabular}{|@{\hskip 1pt}c@{\hskip 1pt}|@{\hskip 1pt}l@{\hskip 2pt}|@{\hskip 1pt}c@{\hskip 1pt}|@{\hskip 1pt}c@{\hskip 1pt}|@{\hskip 1pt}c@{\hskip 1pt}|}
\hline
\textbf{} & \textbf{Filter Step} & \textbf{Stochastic Cloning~(SC)} & \textbf{Forward-time DSKF} & \textbf{Backward-time DSKF} \\ 
\Xhline{1.2pt}
\multirow{5}{*}{\rotatebox[origin=c]{90}{\textbf{\# Multiplies}}} 
& Kalman Gain  & $\ell^2m + \ell m^2 + 2 \ell mn + \ell n^2 $ & $\fDSKFkOneM$  & $\bDSKFkOneM$ \\
&              & $+ m^3 + 2m^2n + mn^2$ & $\fDSKFkTwoM$ & $\bDSKFkTwoM$\\

\cline{2-5}
& State Update & $\ell m + 2mn$ &  $\fDSKFxOneM$ & $\bDSKFxOneM$ \\ 

\cline{2-5}
& Covariance Update~  & $\ell ^2n + \ell mn + 3 \ell n^2 + m^2n $ &  $\fDSKFpOneM$ &  $\bDSKFpOneM$ \\
&                     & $+ 2mn^2 + 2n^3$ & $\fDSKFpTwoM$ &  $\bDSKFpTwoM$ \\
\Xhline{1.2pt}
\multirow{7}{*}{\rotatebox[origin=c]{90}{\textbf{\# Adds}}} 
& Kalman Gain & $\ell ^2m + \ell m^2 + 2\ell mn - \ell m $  &  $\fDSKFkOneA$  & $\bDSKFkOneA$   \\
&             & $+ \ell n^2 - \ell n + m^3 + 2m^2n$ &  $\fDSKFkTwoA$   & $\bDSKFkTwoA$ \\ 
&             & $ + mn^2 - 2mn$ & $\fDSKFkThreeA$ & $\bDSKFkThreeA$ \\

\cline{2-5}
& State Update & $\ell m + 2mn$ & $\fDSKFxOneA$ &  $\bDSKFxOneA$ \\ 
\cline{2-5}
& Covariance Update~ & $\ell^2n + \ell mn + 3\ell n^2 - 2\ell n$ & $\fDSKFpOneA$ & $\bDSKFpOneA$  \\
& & $+ m^2n + 2mn^2 - mn + 2n^3$ & $\fDSKFpTwoA$ & $\bDSKFpTwoA$ \\
& & $- 2n^2 + n$  & $\fDSKFpThreeA$ & $\bDSKFpThreeA$  \\ 
\Xhline{1.2pt}
\multicolumn{2}{|@{\hskip 1pt}c@{\hskip 1pt}|}{\textbf{Total Flops}}
& \begin{tabular}[c]{@{\hskip 4pt}c@{\hskip 4pt}} $2 \ell ^2m + 2 \ell ^2n + 2 \ell m^2 + 6 \ell mn $ \\ $+ \ell m + 8 \ell n^2 - 3 \ell n + 2m^3$ \\ $ + 6m^2n + 6mn^2 + mn + 4n^3 $ \\ 
$- 2n^2 + n$ \\ 
\end{tabular} 


& \begin{tabular}[c]{@{\hskip 4pt}c@{\hskip 4pt}}  $\fDSKFTotOne$ \\
$\fDSKFTotTwo$ \\
$\fDSKFTotThree$ \\
$\fDSKFTotFour$ \\
 \end{tabular} 

& \begin{tabular}[c]{@{\hskip 4pt}c@{\hskip 4pt}}  $\bDSKFTotOne$ \\
$\bDSKFTotTwo$ \\
$\bDSKFTotThree$ \\
$\bDSKFTotFour$ \\
 \end{tabular}  \\

\hline
\end{tabular} \label{tab1}
\end{table}

\newcommand{\scKmemOne}{2 \ell m + 3 \ell n + 2m^2 + 3mn } 
\newcommand{\scKmemTwo}{+ 2m + n^2} 

\newcommand{\scXmem}{\ell + 2m + 2n} 

\newcommand{\scPmem}{2 \ell n + mn + 4n^2 + 1} 

\newcommand{\scTotmemOne}{2 \ell m + 5 \ell n + \ell  + 2m^2 } 
\newcommand{\scTotmemTwo}{+ 4mn + 4m + 5n^2 + 2n} 
\newcommand{\scTotmemThree}{+ 1}

\newcommand{\bdskfKmemOne}{\ell m + 4m^2 + 5mn + 2m }
\newcommand{\bdskfKmemTwo}{+ 2n^2 + 2n}

\newcommand{\bdskfXmem}{3m + n}

\newcommand{\bdskfPmem}{mn + 6n^2 + 1}

\newcommand{\bdskfTotmemOne}{\ell m + 4m^2 + 6mn + 5m}
\newcommand{\bdskfTotmemTwo}{+ 8n^2 + 3n + 1}

\newcommand{\fdskfKmemOne}{\ell m + \ell n + 4m^2 + 6mn}
\newcommand{\fdskfKmemTwo}{ + 2m + n^2}

\newcommand{\fdskfXmem}{3m + n}

\newcommand{\fdskfPmem}{2n^2}

\newcommand{\fdskfTotmemOne}{\ell m + \ell n + 4m^2 + 6mn}
\newcommand{\fdskfTotmemTwo}{+ 5m + 3n^2 + n}

\begin{table}
\begin{center}
\caption{Memory Allocations, in Total Floating-Point Numbers (Floats), for the Delayed-State Filters, with $n$~States, $\ell$~Cloned States, and $m$~Measurements.}\label{tab:memtable}

\begin{tabular}{|c|c|c|c|}\hline
  \textbf{Filter Step} & \textbf{Stochastic Cloning~(SC)} & \textbf{Forward-Time DSKF} & \textbf{Backward-Time DSKF} \\ 
  \hline
  {Kalman Gain}  & $\scKmemOne$ & $\fdskfKmemOne$ & $\bdskfKmemOne$ \\ 
              & $\scKmemTwo$ & $\fdskfKmemTwo$ & $\bdskfKmemTwo$ \\
  \hline

  {State Update} & $\scXmem$ & $\fdskfXmem$ & $\bdskfXmem$ \\
 \hline 

{Covariance Update}  & $\scPmem$ & $\fdskfPmem$ & $\bdskfPmem$\\ 

\Xhline{1.2pt}

\multirow{2}{*}{\textbf{Total}} & $\scTotmemOne$ & $\fdskfTotmemOne$ & $\bdskfTotmemOne$ 
 \\ 
 & $\scTotmemTwo$ & $\fdskfTotmemTwo$ & $\bdskfTotmemTwo$  \\

\hline
\end{tabular}
\end{center}
\end{table}

%% file: sections/10_conclusion.tex
\section{Conclusion}\label{sec:conc}
Stochastic cloning~(SC) is a widely used method for processing delayed-state measurements (such as odometry) in robotics by augmenting the current state with a prior one. 
This approach is oftentimes motivated by the common misperception that Kalman filters, with the state defined at a single time, cannot fully account for cross-time correlations.
This work demonstrates that a properly derived Kalman filter like the lesser-known but well-established delayed-state Kalman filter~(DSKF), yields identical state and covariance updates to those of SC,  without the need for artificial state augmentation.
Additionally, two equivalent formulations of the DSKF are presented, which arise from different but consistent ways of incorporating the required cross-time correlations within the Kalman filter framework.

To be clear, this work does not suggest that one method is better or worse than the others. 
After all, they produce the exact same result under comparable computational complexity and memory usage, and implementation preferences may vary. 
Rather, the intent is to remind the community that state augmentation is not the only way to properly account for correlations, for one only needs to revisit the Kalman filter formulation to incorporate the relevant cross-correlation terms and achieve the same result within a unified framework.
Assertions that Kalman filter variants cannot properly account for correlations are simply not true.

%% file: sections/11_appendix.tex
\section*{Appendix}

The forward-time and backward-time formulations of the DSKF, respectively presented in Sections~\ref{ssec:forwardDSKF}~and~\ref{ssec:backwardDSKF}, are in fact equivalent.
To demonstrate this, the equivalence of the Kalman gains are shown first, followed by the equivalence of the updated state error covariances. 
The backward-time DSKF Kalman gain in Eq.~\eqref{eq:corrKalmanGain} is used as a starting point, and the definitions of $\corr{\linmeasmdl}\tidx{\curidx}$ and $\corr{\crosscov}\tidx{\curidx}$ from Eqs.~\eqref{eq:corrHJDef}~and~\eqref{eq:corrNRDef} are substituted in the first multiplicative term before simplification 
\begin{align}
    \bwdTidx{\kalmangain}{\curidx} 
    &= \left(\pred{\statecov}\tidx{\curidx}\corr{\linmeasmdl}\trans\tidx{\curidx} -  \corr{\crosscov}\trans\tidx{\curidx} \right)
    \left( \corr{\linmeasmdl}\tidx{\curidx}\pred{\statecov}\tidx{\curidx} \corr{\linmeasmdl}\trans\tidx{\curidx} 
    - \corr{\crosscov}\tidx{\curidx} \corr{\linmeasmdl}\trans\tidx{\curidx}
    - \corr{\linmeasmdl}\tidx{\curidx}\corr{\crosscov}\trans\tidx{\curidx}
    + \corr{\measnoisecov}\tidx{\curidx} 
    \right)\inv,
    \\
    &= 
    \left( \pred{\statecov}\tidx{\curidx} \linmeasmdl\tidx{\curidx,\curidx}\trans + \left( \pred{\statecov}\tidx{\curidx} - \procnoisecov\tidx{\curidx} \right) \stm\tidx{\previdx,\curidx}\trans \linmeasmdl\tidx{\previdx,\curidx}\trans \right)
    \left( \corr{\linmeasmdl}\tidx{\curidx}\pred{\statecov}\tidx{\curidx} \corr{\linmeasmdl}\trans\tidx{\curidx} 
    - \corr{\crosscov}\tidx{\curidx} \corr{\linmeasmdl}\trans\tidx{\curidx}
    - \corr{\linmeasmdl}\tidx{\curidx}\corr{\crosscov}\trans\tidx{\curidx}
    + \corr{\measnoisecov}\tidx{\curidx} 
    \right)\inv. \label{eq:kalmanGainEquiv_Mcoeff_1}
\end{align}
After substituting $\pred{\statecov}\tidx{\curidx} - \procnoisecov\tidx{\curidx} = \stm\tidx{\curidx,\previdx} {\statecov}\tidx{\previdx} \stm\tidx{\curidx,\previdx}\trans$ based on Eq.~\eqref{eq:predCov_kf}, a comparison with Eq.~\eqref{eq:finalM} shows that the first multiplicative term reduces to $\Mcoeff\tidx{\curidx}\trans$ 
\begin{align}
    \bwdTidx{\kalmangain}{\curidx} 
    &= \left( \pred{\statecov}\tidx{\curidx} \linmeasmdl\tidx{\curidx,\curidx}\trans + \left( \stm\tidx{\curidx,\previdx} {\statecov}\tidx{\previdx} \stm\tidx{\curidx,\previdx}\trans \right) \stm\tidx{\previdx,\curidx}\trans \linmeasmdl\tidx{\previdx,\curidx}\trans \right)
    \left( \corr{\linmeasmdl}\tidx{\curidx}\pred{\statecov}\tidx{\curidx} \corr{\linmeasmdl}\trans\tidx{\curidx} 
    - \corr{\crosscov}\tidx{\curidx} \corr{\linmeasmdl}\trans\tidx{\curidx}
    - \corr{\linmeasmdl}\tidx{\curidx}\corr{\crosscov}\trans\tidx{\curidx}
    + \corr{\measnoisecov}\tidx{\curidx} 
    \right)\inv,
    \\
    &= \Mcoeff\tidx{\curidx}\trans
    \left( \corr{\linmeasmdl}\tidx{\curidx}\pred{\statecov}\tidx{\curidx} \corr{\linmeasmdl}\trans\tidx{\curidx} 
    - \corr{\crosscov}\tidx{\curidx} \corr{\linmeasmdl}\trans\tidx{\curidx}
    - \corr{\linmeasmdl}\tidx{\curidx}\corr{\crosscov}\trans\tidx{\curidx}
    + \corr{\measnoisecov}\tidx{\curidx} 
    \right)\inv. \label{eq:kalmanGainEquiv_Mcoeff_2}
\end{align} 
The second multiplicative term is similarly expanded by substituting the definitions of $\corr{\linmeasmdl}\tidx{\curidx}$, $\corr{\crosscov}\tidx{\curidx}$, and $\corr{\measnoisecov}\tidx{\curidx}$ from Eqs.~\eqref{eq:corrHJDef}~and~\eqref{eq:corrNRDef}.
Note here that the fully expanded definition of $\corr{\measnoisecov}\tidx{\curidx} = \corr{\linmeasmdlprev}\tidx{\curidx}\procnoisecov\tidx{\curidx}\corr{\linmeasmdlprev}\tidx{\curidx}\trans + \measnoisecov\tidx{\curidx} = {\linmeasmdl}\tidx{\previdx,\curidx} \stm\tidx{\previdx,\curidx} \procnoisecov\tidx{\curidx} \stm\trans\tidx{\previdx,\curidx} {\linmeasmdl}\trans\tidx{\previdx,\curidx} + \measnoisecov\tidx{\curidx}$,
which yields
\begin{align}
    \bwdTidx{\kalmangain}{\curidx} 
    &=
    \Mcoeff\tidx{\curidx}\trans
    \left(
    \linmeasmdl\tidx{\curidx,\curidx} \pred{\statecov}\tidx{\curidx} \linmeasmdl\tidx{\curidx,\curidx}\trans + \measnoisecov\tidx{\curidx} 
    + \linmeasmdl\tidx{\previdx,\curidx}\stm\tidx{\previdx,\curidx} \left(\pred{\statecov}\tidx{\curidx} - \procnoisecov\tidx{\curidx} \right) \linmeasmdl\tidx{\curidx,\curidx}\trans 
    + \linmeasmdl\tidx{\curidx,\curidx} \left(\pred{\statecov}\tidx{\curidx} - \procnoisecov\tidx{\curidx}\right) \stm\trans\tidx{\previdx,\curidx} \linmeasmdl\trans\tidx{\previdx,\curidx}
    \right.
     \nonumber \\ 
    &\qquad  
    \left.
    + \linmeasmdl\tidx{\previdx,\curidx}\stm\tidx{\previdx,\curidx} \left(\pred{\statecov}\tidx{\curidx} - \procnoisecov\tidx{\curidx}\right) \stm\trans\tidx{\previdx,\curidx} \linmeasmdl\trans\tidx{\previdx,\curidx}   
    \right).
\end{align}
By similarly substituting $\pred{\statecov}\tidx{\curidx} - \procnoisecov\tidx{\curidx} = \stm\tidx{\curidx,\previdx} \statecov\tidx{\previdx}\stm\trans\tidx{\curidx,\previdx}$, the argument of the inverse simplifies to $\Lcoeff\tidx{\curidx}$ as defined in Eq.~\eqref{eq:finalL}
\begin{align}
    \bwdTidx{\kalmangain}{\curidx} 
    &= \Mcoeff\tidx{\curidx}\trans
    \left( \linmeasmdl\tidx{\curidx,\curidx} \pred{\statecov}\tidx{\curidx} \linmeasmdl\tidx{\curidx,\curidx}\trans + \measnoisecov\tidx{\curidx} 
    + \linmeasmdl\tidx{\previdx,\curidx}  \statecov\tidx{\previdx}\stm\trans\tidx{\curidx,\previdx}  \linmeasmdl\tidx{\curidx,\curidx}\trans 
    + \linmeasmdl\tidx{\curidx,\curidx} \stm\tidx{\curidx,\previdx} \statecov\tidx{\previdx} \linmeasmdl\trans\tidx{\previdx,\curidx}   
    + \linmeasmdl\tidx{\previdx,\curidx} \statecov\tidx{\previdx} \linmeasmdl\trans\tidx{\previdx,\curidx} \right)\inv,
    \\
    &= \Mcoeff\tidx{\curidx}\trans \Lcoeff\tidx{\curidx}\inv, \label{eq:fwdBwdKalmanGainsEqual}
\end{align}
which is equivalent to the forward-time DSKF Kalman gain $\fwdTidx{\kalmangain}{\curidx}$ expressed in Eq.~\eqref{eq:fwdDSKFkalmanGain}.
This also consequently demonstrates that the state update is equivalent between the forward-time and backward-time DSKF formulations.

Having established the equivalence between the Kalman gains of both DSKF formulations, our attention is now turned to the updated state error covariances.
Starting with the backward-time covariance update from Eq.~\eqref{eq:corrStateCovUpdate} and rearranging terms yields
\begin{align}
    \upd{\statecov}\tidx{\curidx} 
    &= 
    \left( \eye - \bwdTidx{\kalmangain}{\curidx} \corr{\linmeasmdl}\tidx{\curidx} \right) \pred{\statecov}\tidx{\curidx} \left( \eye - \bwdTidx{\kalmangain}{\curidx} \corr{\linmeasmdl}\tidx{\curidx} \right)\trans 
    + \left( \eye - \bwdTidx{\kalmangain}{\curidx} \corr{\linmeasmdl}\tidx{\curidx} \right) \corr{\crosscov}\trans\tidx{\curidx} \bwdTidx{\kalmangain}{\curidx}\trans 
    + \bwdTidx{\kalmangain}{\curidx} \corr{\crosscov}\tidx{\curidx}   \left( \eye - \bwdTidx{\kalmangain}{\curidx} \corr{\linmeasmdl}\tidx{\curidx} \right)\trans
    + \bwdTidx{\kalmangain}{\curidx} \corr{\measnoisecov}\tidx{\curidx} \bwdTidx{\kalmangain}{\curidx}\trans, \\
    &= 
    \pred{\statecov}\tidx{\curidx} 
    - \bwdTidx{\kalmangain}{\curidx} \left( \corr{\linmeasmdl}\tidx{\curidx} \pred{\statecov}\tidx{\curidx} - \corr{\crosscov}\tidx{\curidx} \right) 
    + \left( - \pred{\statecov}\tidx{\curidx} \corr{\linmeasmdl}\tidx{\curidx}\trans \bwdTidx{\kalmangain}{\curidx}\trans 
        +   \bwdTidx{\kalmangain}{\curidx} \corr{\linmeasmdl}\tidx{\curidx}\pred{\statecov}\tidx{\curidx} \corr{\linmeasmdl}\tidx{\curidx}\trans \bwdTidx{\kalmangain}{\curidx}\trans
        + \corr{\crosscov}\trans\tidx{\curidx} \bwdTidx{\kalmangain}{\curidx}\trans
        - \bwdTidx{\kalmangain}{\curidx} \corr{\linmeasmdl}\tidx{\curidx}\corr{\crosscov}\trans\tidx{\curidx} \bwdTidx{\kalmangain}{\curidx}\trans 
        \right. \nonumber \\
    & \qquad \left.
        - \bwdTidx{\kalmangain}{\curidx} \corr{\crosscov}\tidx{\curidx}  \corr{\linmeasmdl}\tidx{\curidx}\trans \bwdTidx{\kalmangain}{\curidx}\trans
        + \bwdTidx{\kalmangain}{\curidx} \corr{\measnoisecov}\tidx{\curidx} \bwdTidx{\kalmangain}{\curidx}\trans
    \right). 
\end{align}
Earlier in Eqs.~\eqref{eq:kalmanGainEquiv_Mcoeff_1}~and~\eqref{eq:kalmanGainEquiv_Mcoeff_2}, it was demonstrated that $\left( \corr{\linmeasmdl}\tidx{\curidx} \pred{\statecov}\tidx{\curidx} - \corr{\crosscov}\tidx{\curidx} \right)=\Mcoeff\tidx{\curidx}\trans$, by substituting the definitions of $\corr{\linmeasmdl}\tidx{\curidx}$ and $\corr{\crosscov}\tidx{\curidx}$ as well as $\pred{\statecov}\tidx{\curidx} - \procnoisecov\tidx{\curidx} = \stm\tidx{\curidx,\previdx} {\statecov}\tidx{\previdx} \stm\tidx{\curidx,\previdx}\trans$.
Now let us substitute in $\Mcoeff\tidx{\curidx}\trans$ and, in the second additive term, use our earlier result from Eq.~\eqref{eq:fwdBwdKalmanGainsEqual} that the forward-time and backward-time DSKF Kalman gains are equivalent
\begin{align}
    \upd{\statecov}\tidx{\curidx} 
    &= 
    \pred{\statecov}\tidx{\curidx} 
    - \fwdTidx{\kalmangain}{\curidx} \Mcoeff\tidx{\curidx}\trans 
    + \left( - \pred{\statecov}\tidx{\curidx} \corr{\linmeasmdl}\tidx{\curidx}\trans \bwdTidx{\kalmangain}{\curidx}\trans 
        +   \bwdTidx{\kalmangain}{\curidx} \corr{\linmeasmdl}\tidx{\curidx}\pred{\statecov}\tidx{\curidx} \corr{\linmeasmdl}\tidx{\curidx}\trans \bwdTidx{\kalmangain}{\curidx}\trans
        + \corr{\crosscov}\trans\tidx{\curidx} \bwdTidx{\kalmangain}{\curidx}\trans
        - \bwdTidx{\kalmangain}{\curidx} \corr{\linmeasmdl}\tidx{\curidx}\corr{\crosscov}\trans\tidx{\curidx} \bwdTidx{\kalmangain}{\curidx}\trans
        \right. \nonumber \\
    & \qquad \left.
        - \bwdTidx{\kalmangain}{\curidx} \corr{\crosscov}\tidx{\curidx}  \corr{\linmeasmdl}\tidx{\curidx}\trans \bwdTidx{\kalmangain}{\curidx}\trans
        + \bwdTidx{\kalmangain}{\curidx} \corr{\measnoisecov}\tidx{\curidx} \bwdTidx{\kalmangain}{\curidx}\trans
    \right). 
\end{align}
Finally, to demonstrate that the updated state error covariances are equivalent in both DSKF formulations, what remains to be shown is that the third additive term goes to zero.
This is achieved by rearranging terms and using the definition of $\bwdTidx{\kalmangain}{\curidx}$ in Eq.~\eqref{eq:corrKalmanGain}
\begin{align}
    \upd{\statecov}\tidx{\curidx} 
    &= \pred{\statecov}\tidx{\curidx} 
    - \fwdTidx{\kalmangain}{\curidx} \Mcoeff\tidx{\curidx}\trans + \left(  \left( -\pred{\statecov}\tidx{\curidx}  \corr{\linmeasmdl}\tidx{\curidx}\trans + \corr{\crosscov}\tidx{\curidx}\trans \right)   \bwdTidx{\kalmangain}{\curidx}\trans 
    + \bwdTidx{\kalmangain}{\curidx} \left( 
        \corr{\linmeasmdl}\tidx{\curidx} \pred{\statecov}\tidx{\curidx}  \corr{\linmeasmdl}\tidx{\curidx}\trans 
        - \corr{\linmeasmdl}\tidx{\curidx}   \corr{\crosscov}\tidx{\curidx}\trans
        - \corr{\crosscov}\tidx{\curidx}   \corr{\linmeasmdl}\tidx{\curidx}\trans
        + \corr{\measnoisecov}\tidx{\curidx} 
        \right) \bwdTidx{\kalmangain}{\curidx}\trans\right), \\
    &= \pred{\statecov}\tidx{\curidx} 
    - \fwdTidx{\kalmangain}{\curidx} \Mcoeff\tidx{\curidx}\trans + \left( \left( -\pred{\statecov}\tidx{\curidx}  \corr{\linmeasmdl}\tidx{\curidx}\trans + \corr{\crosscov}\tidx{\curidx}\trans \right)   \bwdTidx{\kalmangain}{\curidx}\trans \right.
    \nonumber \\
    & \qquad \left.
    + \left(\pred{\statecov}\tidx{\curidx}\corr{\linmeasmdl}\trans\tidx{\curidx} -  \corr{\crosscov}\trans\tidx{\curidx} \right)
    \left( \corr{\linmeasmdl}\tidx{\curidx}\pred{\statecov}\tidx{\curidx} \corr{\linmeasmdl}\trans\tidx{\curidx} 
    - \corr{\crosscov}\tidx{\curidx} \corr{\linmeasmdl}\trans\tidx{\curidx}
    - \corr{\linmeasmdl}\tidx{\curidx}\corr{\crosscov}\trans\tidx{\curidx}
    + \corr{\measnoisecov}\tidx{\curidx} 
    \right)\inv \left( 
        \corr{\linmeasmdl}\tidx{\curidx} \pred{\statecov}\tidx{\curidx}  \corr{\linmeasmdl}\tidx{\curidx}\trans 
        - \corr{\linmeasmdl}\tidx{\curidx}   \corr{\crosscov}\tidx{\curidx}\trans
        - \corr{\crosscov}\tidx{\curidx}   \corr{\linmeasmdl}\tidx{\curidx}\trans
        + \corr{\measnoisecov}\tidx{\curidx} 
        \right) \bwdTidx{\kalmangain}{\curidx}\trans\right), \\
    &= \pred{\statecov}\tidx{\curidx} 
    - \fwdTidx{\kalmangain}{\curidx} \Mcoeff\tidx{\curidx}\trans - \left( \pred{\statecov}\tidx{\curidx}  \corr{\linmeasmdl}\tidx{\curidx}\trans - \corr{\crosscov}\tidx{\curidx}\trans \right)   \bwdTidx{\kalmangain}{\curidx}\trans
    + \left( \pred{\statecov}\tidx{\curidx}  \corr{\linmeasmdl}\tidx{\curidx}\trans - \corr{\crosscov}\tidx{\curidx}\trans \right)   \bwdTidx{\kalmangain}{\curidx}\trans, \\
    &= \pred{\statecov}\tidx{\curidx} 
    - \fwdTidx{\kalmangain}{\curidx} \Mcoeff\tidx{\curidx}\trans,
\end{align}
which is equivalent to the updated state error covariance of the forward-time DSKF formulation in Eq.~\eqref{eq:fwdDSKFStateUpdateCov_final}.

%% file: sections/12_fundingAcknowledgments.tex
\section*{Funding Sources}

This work was funded in part by Georgia Tech Research Institute (GTRI) Independent Research and Development (IRAD) funds.

\section*{Acknowledgments}
The authors would like to thank Christopher Valenta and Frank Dellaert for their feedback on this manuscript. 
During the preparation of this work, ChatGPT~4-5 and Overleaf AI Assist were used to assist with proofreading (e.g., spell checking, grammar) and for suggestions to improve readability. 
The AI's editorial suggestions were reviewed by the authors before incorporation into the final manuscript.

%% file: references.bib
@techreport{carpenter2018navigation,
  title={Navigation Filter Best Practices},
  author={Carpenter, J Russell and D’Souza, Christopher N},
url={https://ntrs.nasa.gov/api/citations/20180003657/downloads/20180003657.pdf},
institution={NASA Engineering and Safety Center},
  year={2018},
  month=apr
}

@inproceedings{roumeliotis2002stochastic,
  title={Stochastic cloning: {A} generalized framework for processing relative state measurements},
  author={Roumeliotis, Stergios I and Burdick, Joel W},
  booktitle={IEEE International Conference on Robotics and Automation},
  year={2002},
  doi = {10.1109/ROBOT.2002.1014801},
  pages={1788--1795}
}

@article{bayard2009reduced,
  title={Reduced-order {K}alman filtering with relative measurements},
  author={Bayard, David S},
  journal={Journal of guidance, control, and dynamics},
  volume={32},
  number={2},
  pages={679--686},
  year={2009}
}

@book{brown1997kalman,
  author       = {Brown, Robert Grover and Hwang, Patrick Y. C.},
  title        = {Introduction to Random Signals and Applied {K}alman Filtering},
  edition      = {3rd},
  publisher    = {John Wiley \& Sons},
  year         = {1997},
  pages        = {348--353}
}

@article{Christian:2021vo,
  title={Image-Based Lunar Terrain Relative Navigation without a Map: Measurements},
  author={J. A. Christian and L. Hong and P. {McKee} and R. Christensen and T. P. Crain},
  journal={Journal of Spacecraft and Rockets},
  volume={58},
  number={1},
  pages={164--181},
  year={2021},
  doi = {10.2514/1.A34875}
}

@book{christian2025optical,
  author    = {John A. Christian},
  title     = {Fundamentals of Spacecraft Optical Navigation},
  year      = {2025},
  publisher = {Wiley},
  address   = {Hoboken, NJ},
  isbn      = {9781394267712}
}

@inproceedings{Bayard:2019,
  title={Vision-Based Navigation for the {NASA} {M}ars Helicopter},
  author={D. S. Bayard and D. T. Conway and R. Brockers and J. H. Delaune and L. H. Matthies and H. F. Grip and G. B. Merewether and T. L. Brown and A. M. {San Martin}},
  Booktitle = {{AIAA SciTech Forum}},
  year={2019},
  pages={1-22},
  doi = {10.2514/6.2019-1411}
}

@inproceedings{Nister:2004,
  title={Visual odometry},
  author={D. Nist{\'e}r and O. Naroditsky and J. Bergen},
  booktitle={IEEE Computer Society Conference on Computer Vision and Pattern Recognition (CVPR).},
  year={2004},
  pages={1-8},
  doi = {10.1109/CVPR.2004.1315094}
}

@article{Scaramuzza:2011,
	Author = {D. Scaramuzza and F. Fraundorfer},
	Journal = {{IEEE Robotics and Automation Magazine}},
	Volume = {18},
	Number = {4},
	Pages = {80--92},
	Title = {Visual Odometry},
	Year = {2011},
	doi = {10.1109/MRA.2011.943233}
}

@article{Forster:2017,
	Author = {C. Forster and L. Carlone and F. Dellaert and D. Scaramuzza},
	Journal = {{IEEE Transactions on Robotics}},
	Volume = {33},
	Number = {1},
	Title = {On-Manifold Preintegration for Real-Time Visual-Inertial Odometry},
	Year = {2017},
	doi = {10.1109/TRO.2016.2597321}
}

@book{Maybeck:1982,
author = {P.S. Maybeck},
title = {Stochastic Models: Estimation and Control: Volume 2},
year = {1982},
Publisher = {Academic Press},
address = {New York, NY},
}

@article{Johnson:2023,
    author = {A. E. Johnson and Y. Cheng and N. Trawny and J. F. Montgomery and S. Schroeder and J. Chang and D. Clouse and S. Aaron and S. Mohan},
    title = {{Implementation of a Map Relative Localization System for Planetary Landing}},
    year = {2023},
    volume = {46},
    issue = {4},
    journal ={{Journal of Guidance, Control, and Dynamics}},
    pages = {618--637},
    doi = {10.2514/1.G006780}
}

@article{xu2021fast,
  title={{FAST-LIO}: A fast, robust {LiDAR}-inertial odometry package by tightly-coupled iterated {K}alman filter},
  author={Xu, Wei and Zhang, Fu},
  journal={IEEE Robotics and Automation Letters},
  volume={6},
  number={2},
  pages={3317--3324},
  year={2021},
  publisher={IEEE}
}

@inproceedings{shan2020lio,
  title={{LIO-SAM}: Tightly-coupled lidar inertial odometry via smoothing and mapping},
  author={Shan, Tixiao and Englot, Brendan and Meyers, Drew and Wang, Wei and Ratti, Carlo and Rus, Daniela},
  booktitle={2020 IEEE/RSJ international conference on intelligent robots and systems (IROS)},
  pages={5135--5142},
  year={2020},
  organization={IEEE}
}

@book{thrun2005probabilistic,
  title={Probabilistic robotics},
  author={Thrun, Sebastian and Burgard, Wolfram and Fox, Dieter},
  publisher={MIT Press},
  address={Cambridge, MA},
  year={2005},
  isbn={978-0-262-20162-9}
}

@inproceedings{mourikis2007multi,
  title={A multi-state constraint {K}alman filter for vision-aided inertial navigation},
  author={Mourikis, Anastasios I and Roumeliotis, Stergios I},
  booktitle={Proceedings 2007 IEEE international conference on robotics and automation},
  pages={3565--3572},
  year={2007},
  organization={IEEE}
}

@article{scaramuzza2019visual,
  title={Visual-inertial odometry of aerial robots},
  author={Scaramuzza, Davide and Zhang, Zichao},
  journal={arXiv preprint arXiv:1906.03289},
  year={2019}
}

@article{nister2006visual,
  title={Visual odometry for ground vehicle applications},
  author={Nist{\'e}r, David and Naroditsky, Oleg and Bergen, James},
  journal={Journal of Field Robotics},
  volume={23},
  number={1},
  pages={3--20},
  year={2006},
  publisher={Wiley Online Library}
}

@article{mohamed2019survey,
  title={A survey on odometry for autonomous navigation systems},
  author={Mohamed, Sherif AS and Haghbayan, Mohammad-Hashem and Westerlund, Tomi and Heikkonen, Jukka and Tenhunen, Hannu and Plosila, Juha},
  journal={IEEE access},
  volume={7},
  pages={97466--97486},
  year={2019},
  publisher={IEEE}
}

@inproceedings{cheng2005visual,
  title={Visual odometry on the {M}ars exploration rovers},
  author={Cheng, Yang and Maimone, Mark and Matthies, Larry},
  booktitle={2005 IEEE International Conference on Systems, Man and Cybernetics},
  volume={1},
  pages={903--910},
  year={2005},
  organization={IEEE}
}

@inproceedings{brown1987proper,
  title={Proper Treatment of the Delta-Range Measurement in an Integrated {GPS}/Inertial System},
  author={Brown, R Grover and McBurney, Paul W},
  booktitle={Proceedings of the 1987 National Technical Meeting of The Institute of Navigation},
  pages={32--39},
  year={1987}
}

@book{kaplan2017understanding,
  title={Understanding {GPS}/{GNSS}: Principles and applications},
  author={Kaplan, Elliott D and Hegarty, Christopher},
  year={2017},
  publisher={Artech house}
}

@inproceedings{guillard2022using,
  title={Using {TDCP} Measurements in a low-cost {PPP}-{IMU} hybridized filter for real-time applications},
  author={Guillard, Anthony and Thevenon, Paul and Milner, Carl},
  booktitle={Proceedings of the 35th International Technical Meeting of the Satellite Division of The Institute of Navigation (ION GNSS+ 2022)},
  pages={2090--2103},
  year={2022}
}

@article{chiang2020design,
  title={The design a {TDCP}-smoothed {GNSS}/odometer integration scheme with vehicular-motion constraint and robust regression},
  author={Chiang, Kai-Wei and Li, Yu-Hua and Hsu, Li-Ta and Chu, Feng-Yu},
  journal={Remote Sensing},
  volume={12},
  number={16},
  pages={2550},
  year={2020},
  publisher={MDPI}
}

@inproceedings{iiyama2023terrestrial,
  title={Terrestrial {GPS} time-differenced carrier-phase positioning of lunar surface users},
  author={Iiyama, Keidai and Bhamidipati, Sriramya and Gao, Grace},
  booktitle={2023 IEEE Aerospace Conference},
  pages={1--9},
  year={2023},
  organization={IEEE}
}

@article{gopalakrishnan2011incorporating,
  title={Incorporating delayed and infrequent measurements in Extended {K}alman Filter based nonlinear state estimation},
  author={Gopalakrishnan, Ajit and Kaisare, Niket S and Narasimhan, Shankar},
  journal={Journal of Process Control},
  volume={21},
  number={1},
  pages={119--129},
  year={2011},
  publisher={Elsevier}
}

@article{oskiper2015augmented,
  title={Augmented reality binoculars},
  author={Oskiper, Taragay and Sizintsev, Mikhail and Branzoi, Vlad and Samarasekera, Supun and Kumar, Rakesh},
  journal={IEEE transactions on visualization and computer graphics},
  volume={21},
  number={5},
  pages={611--623},
  year={2015},
  publisher={IEEE}
}

@inproceedings{schwarze2015intuitive,
  title={An intuitive mobility aid for visually impaired people based on stereo vision},
  author={Schwarze, Tobias and Lauer, Martin and Schwaab, Manuel and Romanovas, Michailas and Bohm, Sandra and Jurgensohn, Thomas},
  booktitle={Proceedings of the IEEE International Conference on Computer Vision Workshops},
  pages={17--25},
  year={2015}
}

@inproceedings{ruppelt2015novel,
  title={A novel finite state machine based step detection technique for pedestrian navigation systems},
  author={Ruppelt, Jan and Kronenwett, Nikolai and Trommer, Gert F},
  booktitle={2015 International Conference on Indoor Positioning and Indoor Navigation (IPIN)},
  pages={1--7},
  year={2015},
  organization={IEEE}
}

@article{mourikis2007sc,
  title={SC-KF mobile robot localization: A stochastic cloning Kalman filter for processing relative-state measurements},
  author={Mourikis, Anastasios I and Roumeliotis, Stergios I and Burdick, Joel W},
  journal={IEEE Transactions on Robotics},
  volume={23},
  number={4},
  pages={717--730},
  year={2007},
  publisher={IEEE}
}

@inproceedings{das2019ekf,
  title={{AS-EKF}: A delay aware state estimation technique for telepresence robot navigation},
  author={Das, Barnali and Dobie, Gordon and Pierce, Stephen Gareth},
  booktitle={2019 Third IEEE International Conference on Robotic Computing (IRC)},
  pages={624--629},
  year={2019},
  organization={IEEE}
}

@inproceedings{merwe2004sigma,
  title={Sigma-point {K}alman filters for integrated navigation},
  author={Merwe, Rudolph van der and Wan, Eric A},
  booktitle={Proceedings of the 60th annual meeting of the institute of navigation (2004)},
  pages={641--654},
  year={2004}
}

@article{liang2021novel,
  title={A novel inertial-aided visible light positioning system using modulated {LED}s and unmodulated lights as landmarks},
  author={Liang, Qing and Sun, Yuxiang and Wang, Lujia and Liu, Ming},
  journal={IEEE Transactions on Automation Science and Engineering},
  volume={19},
  number={4},
  pages={3049--3067},
  year={2021},
  publisher={IEEE}
}

@inproceedings{welte2019four,
  title={Four-wheeled dead-reckoning model calibration using {RTS} smoothing},
  author={Welte, Anthony and Xu, Philippe and Bonnifait, Philippe},
  booktitle={2019 International conference on robotics and automation (ICRA)},
  pages={312--318},
  year={2019},
  organization={IEEE}
}

@inproceedings{helmick2004path,
  title={Path following using visual odometry for a {M}ars rover in high-slip environments},
  author={Helmick, Daniel M and Cheng, Yang and Clouse, Daniel S and Matthies, Larry H and Roumeliotis, Stergios I},
  booktitle={2004 IEEE Aerospace Conference Proceedings (IEEE Cat. No. 04TH8720)},
  volume={2},
  pages={772--789},
  year={2004},
  organization={IEEE}
}

@inproceedings{chenavier1992position,
  title={Position estimation for a mobile robot using vision and odometry.},
  author={Chenavier, Fr{\'e}d{\'e}ric and Crowley, James L},
  booktitle={ICRA},
  volume={89},
  pages={2588--2593},
  year={1992}
}

@article{borenstein1996measurement,
  title={Measurement and correction of systematic odometry errors in mobile robots},
  author={Borenstein, Johann and Feng, Liqiang},
  journal={IEEE Transactions on robotics and automation},
  volume={12},
  number={6},
  pages={869--880},
  year={1996},
  publisher={IEEE}
}

@article{cheng2004mars,
  title={The {M}ars exploration rovers descent image motion estimation system},
  author={Cheng, Yang and Goguen, Jay and Johnson, Andrew and Leger, Chris and Matthies, Larry and Martin, Miguel San and Willson, Reg},
  journal={IEEE Intelligent Systems},
  volume={19},
  number={3},
  pages={13--21},
  year={2004},
  publisher={IEEE}
}

@inproceedings{larsen1998incorporation,
  title={Incorporation of time delayed measurements in a discrete-time {K}alman filter},
  author={Larsen, Thomas Dall and Andersen, Nils A and Ravn, Ole and Poulsen, Niels Kj{\o}lstad},
  booktitle={Proceedings of the 37th IEEE Conference on Decision and Control (Cat. No. 98CH36171)},
  volume={4},
  pages={3972--3977},
  year={1998},
  organization={IEEE}
}

@book{hajek2015random,
  title = {Random Processes for Engineers},
  author = {Hajek, Bruce},
  year = {2015},
  publisher = {Cambridge University Press}
}

@inproceedings{brown1968kalman,
  title={{K}alman filter with delayed states as observables},
  author={Brown, RG and Hartman, GL},
  booktitle={Proceedings of the National Electronics Conference},
  volume={24},
  pages={67--72},
  year={1968}
}

@article{silvestrini2022optical,
  title={Optical navigation for Lunar landing based on Convolutional Neural Network crater detector},
  author={Silvestrini, Stefano and Piccinin, Margherita and Zanotti, Giovanni and Brandonisio, Andrea and Bloise, Ilaria and Feruglio, Lorenzo and Lunghi, Paolo and Lavagna, Mich{\`e}le and Varile, Mattia},
  journal={Aerospace Science and Technology},
  volume={123},
  pages={107503},
  year={2022},
  publisher={Elsevier}
}

@inproceedings{ranganathan2007fast,
  title={Fast {3D} pose estimation with out-of-sequence measurements},
  author={Ranganathan, Ananth and Kaess, Michael and Dellaert, Frank},
  booktitle={2007 IEEE/RSJ International Conference on Intelligent Robots and Systems},
  pages={2486--2493},
  year={2007},
  organization={IEEE}
}

@article{rauch1963solutions,
  title={Solutions to the linear smoothing problem},
  author={Rauch, H},
  journal={IEEE Transactions on Automatic Control},
  volume={8},
  number={4},
  pages={371--372},
  year={1963},
  publisher={IEEE}
}

@article{challa2002fixed,
  title={A fixed-lag smoothing solution to out-of-sequence information fusion problems},
  author={Challa, Subhash and Evans, Robin J and Wang, Xuezhi and Legg, Jonathan},
  journal={Communications in Information and Systems},
  volume={2},
  number={4},
  pages={325--348},
  year={2002},
  publisher={Citeseer}
}

@phdthesis{finch1970smoothing,
  title     = {Smoothing for Delayed State Model with Applications to Aided Inertial Navigation Systems},
  author    = {Finch, Robert Bruce},
  school    = {Iowa State University},
  year      = {1970},
  type      = {{P}h.{D}. dissertation},
  address = {Ames, Iowa}
}

@BOOK{laug,
      AUTHOR = {Anderson, E. and Bai, Z. and Bischof, C. and
                Blackford, S. and Demmel, J. and Dongarra, J. and
                Du Croz, J. and Greenbaum, A. and Hammarling, S. and
                McKenney, A. and Sorensen, D.},
      TITLE = {{LAPACK} Users' Guide},
      EDITION = {Third},
      PUBLISHER = {Society for Industrial and Applied Mathematics},
      YEAR = {1999},
      ADDRESS = {Philadelphia, PA},
      ISBN = {0-89871-447-8 (paperback)} }

@BOOK{fortran,
      AUTHOR = {Press, W. and Teukolsky, S. and Vetterling, W. and Flannery, B.},
      TITLE = {Numerical Recipes in FORTRAN 77},
      EDITION = {Second},
      PUBLISHER = {University of Cambridge},
      YEAR = {1992},
      }
